\pdfoutput=1

\documentclass[11pt]{article}

\usepackage[]{emnlp2021}

\usepackage{times}
\usepackage{latexsym}

\usepackage{url}
\usepackage{times}
\usepackage{amssymb} 
\usepackage{microtype}
\usepackage{graphicx}
\usepackage{booktabs}
\usepackage{floatrow}
\usepackage{booktabs}
\usepackage{numprint}
\usepackage{subcaption} 
\usepackage{enumerate}
\usepackage{todonotes}
\usepackage{bm}
\usepackage{multirow}
\usepackage{arydshln}
\usepackage{xcolor,soul}
\usepackage{amssymb}
\usepackage{pifont}
\usepackage{xurl}
\usepackage{pifont}
\usepackage{amsmath}
\usepackage{arydshln}

\newcommand*\colourcheck[1]{%
  \expandafter\newcommand\csname #1check\endcsname{\textcolor{#1}{\ding{52}}}%
}
\newcommand*\colourcross[1]{%
  \expandafter\newcommand\csname #1cross\endcsname{\textcolor{#1}{\ding{55}}}%
}
\colourcheck{blue}
\colourcheck{green}
\colourcheck{red}

\colourcross{blue}
\colourcross{green}
\colourcross{red}

\usepackage{tikz}
\newcommand*\circled[1]{\tikz[baseline=(char.base)]{
            \node[shape=circle,draw,inner sep=0.8pt] (char) {#1};}}

\newcommand{\rmbf}[1]{\textrm{\textbf{#1}}}

\usepackage[T1]{fontenc}

\usepackage[utf8]{inputenc}

\usepackage{microtype}

\title{Hyper-X: A Unified Hypernetwork for Multi-Task Multilingual Transfer}

\author{Ahmet \"{U}st\"{u}n{\normalfont\textsuperscript{1}},
~Arianna Bisazza{\normalfont\textsuperscript{1}},
~Gosse Bouma{\normalfont\textsuperscript{1}}, \\
\textbf{Gertjan van Noord{\normalfont\textsuperscript{1}},
~Sebastian Ruder{\normalfont\textsuperscript{2}}} \vspace{.1cm}\\
{\normalfont\textsuperscript{1}}University of Groningen \\
{\normalfont\textsuperscript{2}}Google Research\\
\texttt{a.ustun@rug.nl} \\}

\begin{document}
\maketitle
\begin{abstract}
Massively multilingual models are promising for transfer learning across tasks and languages. However, existing methods are unable to fully leverage training data when it is available in different task-language combinations. To exploit such heterogeneous supervision, we propose \textbf{Hyper-X}, a single hypernetwork that unifies multi-task and multilingual learning with efficient adaptation. This model generates weights for adapter modules conditioned on both tasks and language embeddings. By learning to combine task and language-specific knowledge, our model enables zero-shot transfer for unseen languages and task-language combinations. Our experiments on a diverse set of languages demonstrate that Hyper-X achieves the best or competitive gain when a mixture of multiple resources is available, while being on par with strong baselines in the standard scenario. Hyper-X is also considerably more efficient in terms of parameters and resources compared to methods that train separate adapters. Finally, Hyper-X consistently produces strong results in few-shot scenarios for new languages, showing the versatility of our approach beyond zero-shot transfer.\footnote{Our code for Hyper-X will be released at\\ \url{https://github.com/ahmetustun/hyperx}} 
\end{abstract}

\section{Introduction}

\begin{figure}[t]
  \begin{subfigure}[b]{\textwidth} 
    \includegraphics[scale=0.48]{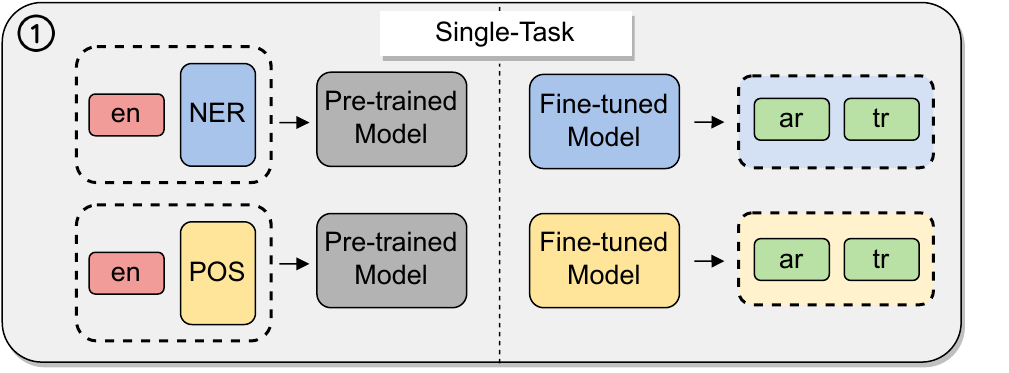}
  \end{subfigure}
  \begin{subfigure}[b]{\textwidth} 
    \includegraphics[scale=0.48]{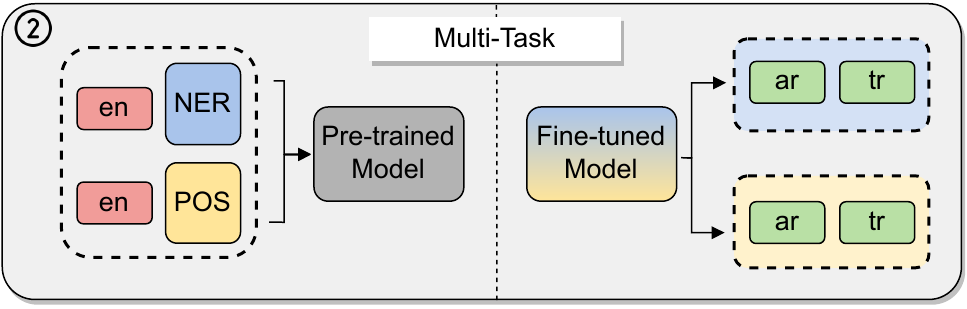} 
  \end{subfigure}
  \begin{subfigure}[b]{\textwidth} 
    \includegraphics[scale=0.48]{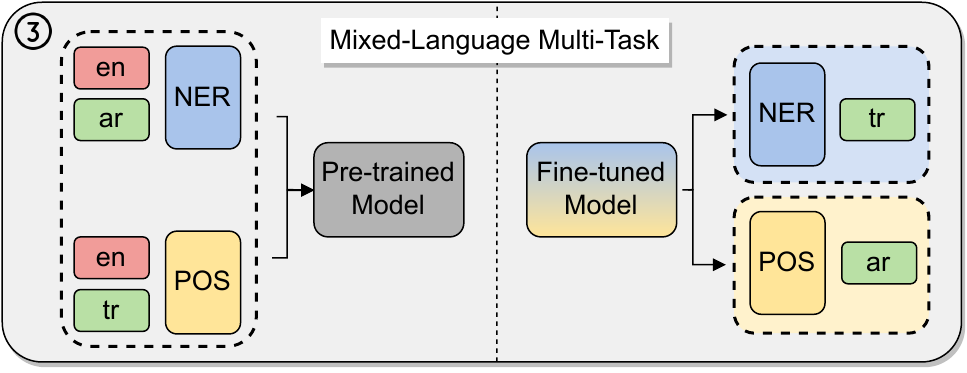}
  \end{subfigure}
  \caption{Experimental settings of different (zero-shot) cross-lingual transfer scenarios. Single-task (1) is the standard setting; multi-task (2) enables cross-task transfer. Mixed-language multi-task (3) additionally allows leveraging task data from multiple source languages for different tasks.}
  \label{fig:exp-setup}
\end{figure}

Transfer learning across languages and tasks 
has long been an important focus in NLP \cite{ruder2019transfer}. Recent advances in massively multilingual transformers \citep[MMTs;][]{devlin2018bert,conneau2019unsupervised} show great success in this area. A benefit of such models is their ability to transfer task-specific information in a high-resource source language to a low-resource target language (Figure \ref{fig:exp-setup}, \circled{1}).
Alternatively, such models can leverage knowledge from multiple tasks for potentially stronger generalization (Figure \ref{fig:exp-setup}, \circled{2}).

Over time, many research communities have been developing resources for specific languages of focus \cite{strassel2016lorelei,11234/1-2895,wilie-etal-2020-indonlu}. In practice, it is thus common for data to be available for different tasks in a mixture of different languages. For instance, in addition to English data for both POS tagging and Named Entity Recognition (NER), a treebank with POS annotation may be available for Turkish, while NER data may be available for Arabic. This example is illustrated in Figure~\ref{fig:exp-setup}, \circled{3}.

\begin{table*}[t]
\small
\centering
\setlength{\tabcolsep}{4pt}
\begin{tabular}{@{}llccccc@{}}
\toprule
\textsc{Model} & \textsc{Description} & X-\textbf{L}ang. & New Lang. & M-\textbf{T}ask & X-Pair (\textbf{LT}) \\ \midrule
MAD-X  & \multirow{2}{*}{Cross-lingual transfer via language/task adapters} & \multirow{2}{*}{\bluecheck} & \multirow{2}{*}{\bluecheck} & \multirow{2}{*}{\redcross} & \multirow{2}{*}{\bluecheck} \\ 
\cite{pfeiffer2020mad} \\ [0.1cm]
HyperFormer & \multirow{2}{*}{Multi-task learning via shared hypernet adapters} & \multirow{2}{*}{\redcross} & \multirow{2}{*}{\redcross} & \multirow{2}{*}{\bluecheck} & \multirow{2}{*}{\redcross} \\
\cite{karimi-mahabadi-etal-2021-parameter} & & \\ [0.1cm]
Parameter Space Fact.  & \multirow{2}{*}{Transfer to unseen task-language pairs via PSF} & \multirow{2}{*}{\redcross} & \multirow{2}{*}{\redcross} & \multirow{2}{*}{\bluecheck} & \multirow{2}{*}{\bluecheck} \\ 
\cite[PSF;][]{ponti-etal-2021-parameter} \\  [0.15cm]
\hdashline 
\noalign{\smallskip} 
\noalign{\smallskip} 
Hyper-X (this work) & Multi-language/task transfer via a unified hypernet & \bluecheck & \bluecheck & \bluecheck & \bluecheck \\
\bottomrule
\end{tabular}
\caption{A comparison of existing approaches and Hyper-X based on their transfer capabilities. We characterize approaches based on whether they can perform cross-lingual transfer (X-Lang.) and cross-task transfer via multi-task learning (M-Task) in the zero-shot setting or to unseen language-task pairs (X-Pair). As a particular case of cross-lingual transfer, `New Lang' represents the case when transfer is generalizable to unseen languages not covered by the multilingual pre-trained model.} 
\label{tab:intro}
\end{table*}

In contrast to existing cross-lingual transfer paradigms such as single-task zero-shot transfer \cite{Hu2020xtreme} or few-shot learning \cite{Lauscher2020}, multi-task learning on such a mixture of datasets (mixed-language multi-task) poses an opportunity to leverage all available data and to transfer information across both tasks and languages to unseen task--language combinations ~\citep{ponti-etal-2021-parameter}.

Standard fine-tuning strategies, however, are limited in their ability to leverage such heterogeneous task and language data. Specifically, MMTs are prone to suffer from catastrophic forgetting and interference \citep{wang-etal-2020-negative} when they are fine-tuned on multiple sources. Adapters \citep{houlsby2019parameter}, a parameter-efficient fine-tuning alternative are commonly used for transfer either across tasks \citep{karimi-mahabadi-etal-2021-parameter} or languages \citep{ustun-etal-2020-udapter} but require training a new adapter for each new language \citep{pfeiffer2020mad}. 


In this paper, we propose a unified hypernetwork, \textbf{\textsc{Hyper-X}} that is particularly suited to this setting by leveraging multiple sources of information including different languages and tasks within a single model. The core idea consists of taking language and task embeddings as input, and generating adapter parameters via a hypernetwork for the corresponding task-language combination.
By parameterizing each task and language separately, Hyper-X enables adaptation to unseen combinations at test time while exploiting all available data resources. 

Additionally, Hyper-X can make seamless use of masked language modelling (MLM) on unlabelled data, which enables it to perform zero-shot adaptation to languages not covered by the MMT during pre-training. MLM also enables Hyper-X to learn a language representation even without available task-specific data. 

In sum, our work brings together a number of successful transfer `ingredients' that have been explored in very recent literature (see Table~\ref{tab:intro}), namely multi-task learning, multilingual learning, further pre-training, along a high degree of compute- and time-efficiency.

We evaluate Hyper-X for cross-lingual transfer on two sequence labelling tasks, namely part-of-speech (POS) tagging and named-entity recognition (NER) in 16 languages---7 of which are not covered in pre-training---across the three experimental setups depicted in Figure~\ref{fig:exp-setup}.
Our experiments demonstrate that Hyper-X is on par with strong baselines for cross-lingual transfer from English. In the multi-task and mixed-language settings, Hyper-X shows a large improvement compared to the standard baselines and matches the performance of the less efficient adapter-based model due to its ability to leverage heterogeneous sources of supervision. 
Analysis highlights that Hyper-X is superior in terms of efficiency--performance trade-offs. Finally, we evaluate our model in a few-shot setting, where Hyper-X consistently achieves competitive performance across different languages and tasks, which suggests the usability of our approach in continuous learning scenarios. 

\section{Background}


\subsection{Adapters}
Adapters \citep{rebuffi2018efficient}
are light-weight bottleneck layers inserted into a MMT to fine-tune the model for a new task \citep{houlsby2019parameter}, language \citep{pfeiffer2020mad} or domain \citep{bapna-firat-2019-simple}. The pre-trained weights of the transformer remain fixed and only adapter parameters are updated. This setup 
prevents catastrophic forgetting \citep{mccloskey1989catastrophic} by encapsulating specialized knowledge.

Formally, an adapter module $\textrm{A}_{i}$ at layer $i$ consists of a down-projection $\rmbf{D}_i\in \mathbb{R}^{h\times b}$ of the input $\rmbf{z}_i\in \mathbb{R}^h$ with the bottleneck dimension $b$, a non-linear function (ReLU) and an up-projection $\rmbf{U}_i\in \mathbb{R}^{b\times h}$ :
\begin{equation}
\textrm{A}_{i}(\rmbf{z}_i) = \rmbf{U}_i.\textrm{ReLU}(\rmbf{D}_i.\rmbf{z}_i) + \rmbf{z}_i
\end{equation}
\noindent where this feed-forward network is followed by a residual link connecting to the input $\rmbf{z}_i$.

\begin{figure}[t]
    \centering \hspace{-0.4cm}
    \includegraphics[scale=0.58]{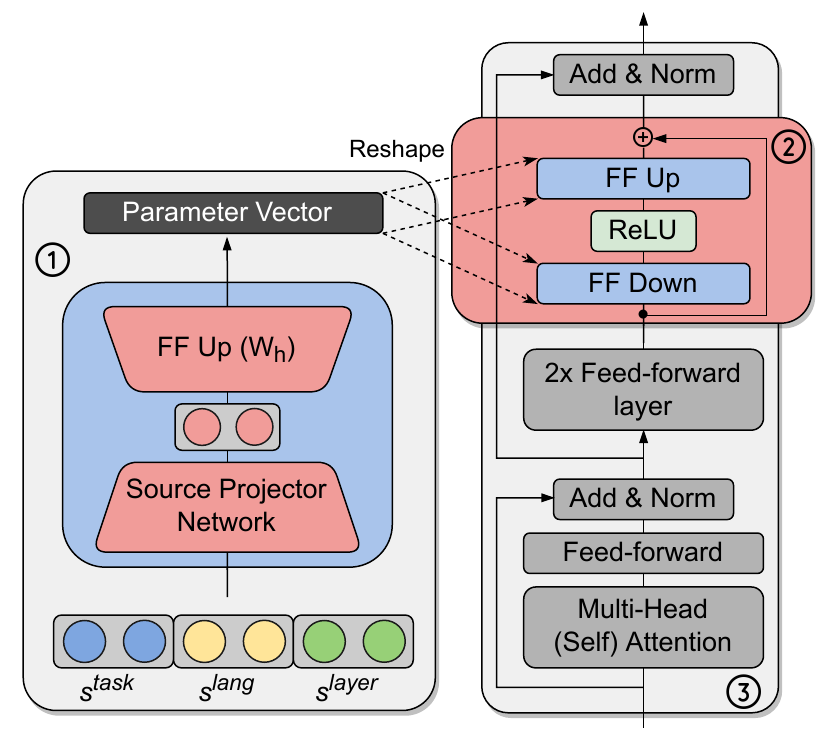}
    \caption{Overview of Hyper-X. The hypernetwork (1) takes the concatenation of task, language and layer embeddings as input and generates a flat parameter vector. Before the final transformation, the source projector network projects the combination of these embeddings to a smaller dimension. The parameter vector is then reshaped and cast to weights of the adapter (2), which are inserted into a transformer layer (3).}
  \label{fig:hyperx}
\end{figure}

\subsection{Hypernetworks}

A hypernetwork is a network that generates the weights for a larger main network \citep{ha2016hypernetworks}. When using a hypernetwork, the main model learns the desired objective (e.g.\ classification) whereas the hypernetwork takes an auxiliary input (usually an embedding) that represents the structure of the weights and generates parameters of the main model. A hypernetwork thus enables learning a single parameter space shared across multiple transfer dimensions 
such as tasks \citep{karimi-mahabadi-etal-2021-parameter} or languages \citep{platanios2018contextual} while also allowing input-specific reparametrization.

More concretely, a hypernetwork is a generator function $\mathcal{H}$ that takes an embedding $\rmbf{s}^{(h)}\in \mathbb{R}^{d_s}$ representing the input sources, 
and generates the model parameters $\Theta$:
\begin{equation}
    \Theta \triangleq \mathcal{H}(\rmbf{s}^{(h)})
\end{equation}
\noindent While $\mathcal{H}$ can be any differentiable function, it is commonly parameterized as a simple linear transform ($\rmbf{W}_h$) that generates a flat vector with the dimension of $d_a$, which corresponds to the total number of model parameters. $\rmbf{W}_h$ is shared across all input sources, enabling maximum sharing. 

\section{Hyper-X}

We propose, Hyper-X, an efficient adaptation of a MMT by exploiting multiple sources of information
for transfer to an unseen language or task-language pairs. Specifically, Hyper-X learns to combine task and language-specific knowledge in the form of embeddings using a hypernetwork. Conditioned on the task and language embeddings, the hypernetwork generates \textit{composite} adapter layers for the corresponding task-language combination (e.g. NER in Turkish), thereby enabling transfer to arbitrary task-language pairs at test time. 
Figure~\ref{fig:hyperx} provides an overview of our model. 

By jointly learning from task and language information, Hyper-X overcomes some of the limitations of prior work: Unlike adapter-based approaches \citep{pfeiffer2020mad,ustun-etal-2020-udapter} that transfer cross-lingual information only to the task of the task adapter, our model is capable of leveraging supervision---and positive transfer---from both multiple tasks and languages. Moreover, unlike \citet{ponti-etal-2021-parameter} who require annotated data in one of the target tasks for each language, Hyper-X is able to perform zero-shot transfer even when there is no annotated data from any of the target tasks, by using MLM as an auxiliary task for each language. 

\subsection{A Hypernetwork for Task-Language Adapters}
\label{sec:hypernet}

We use a standard hypernetwork as the parameter generator function. However, instead of generating the full model parameters, our hypernetwork generates the parameters for each adapter layer. Concretely, the hypernetwork $\mathcal{H}$ generates adapter parameters where each adapter layer $A_i$ consists of down and up-projection matrices ($\rmbf{D}_i$, $\rmbf{U}_i$):
\begin{equation}
    \rmbf{D}_i, \rmbf{U}_i \triangleq \mathcal{H}(\rmbf{s}^{(h)})
\end{equation}

\paragraph{Decoupling Tasks and Languages} In Hyper-X, we condition the parameter generation on the input task and language. Therefore, given a combination of task $t\in \{t_1,...,t_m\}$ and language $l\in \{l_1,...,l_n\}$, the source embedding contains knowledge from both sources: $\rmbf{s}^{(h)} \approx (t,l)$. We parameterize each task and language via separate embeddings, which enables adaptation to any task-language combination. Task and language embeddings ($\rmbf{s}^{(t)},\rmbf{s}^{(l)}$) are low-dimensional vectors that are learned together with the parameters of the hypernetwork. During training, for each mini-batch we update these embeddings according to the task and language that the mini-batch is sampled from.

\paragraph{MLM as Auxiliary Task} Hyper-X learns separate tasks and languages 
embeddings---as long as the task and language have been seen during training. As annotated data in many under-represented languages is limited, we employ MLM as an auxiliary task during training to enable computing embeddings for every language. Moreover, MLM enables a better zero-shot performance for languages that are not included in MMT pre-training (see \S~\ref{sec:mlm} for a detailed analysis of the impact of MLM).  

\paragraph{Sharing Across Layers} In addition to the task and language embedding, we learn a layer embedding $\rmbf{s}^{(i)}$ \citep{karimi-mahabadi-etal-2021-parameter,ansell-etal-2021-mad-g} corresponding to the transformer layer index $i$ where the respective adapter module is plugged in. Since Hyper-X generates an adapter for each Transformer layer, learning independent layer embeddings allows for information sharing across those layers. Moreover, as layer embeddings allow the use of a single hypernetwork for all Transformer layers, they reduce the trainable parameters, i.e., size of the hypernetwork, by a factor corresponding to the number of layers of the main model.

\paragraph{Combining Multiple Sources} To combine language, task and layer embeddings,
we use a simple source projector network $\mathcal{P}_s$ as part of our hypernetwork. This module consisting of two feed-forward layers with a ReLU activation takes the concatenation of the three embeddings and learns a combined embedding $\rmbf{s}^{(p)} \in \mathbb{R}^{d_p}$ with a potentially smaller dimension:
\begin{eqnarray}
    \rmbf{s}^{(h)} &=& \rmbf{s}^{(l)}\oplus~\rmbf{s}^{(t)}\oplus~\rmbf{s}^{(i)} \\ 
    \rmbf{s}^{(p)} &=& \mathcal{P}_s(\rmbf{s}^{(h)})
\end{eqnarray}

\noindent where $\rmbf{s}^{(h)} \in \mathbb{R}^{d_s}$ refers to the concatenated embedding before the $\mathcal{P}_s$, with $d_s = d_l+d_t+d_i$. This component enables learning how to combine source embeddings while also reducing the total number of trainable parameters. 

\section{Experiments}
\label{sec:experiments}

\paragraph{Dataset and Languages} We conduct experiments on two downstream tasks: part-of-speech (POS) tagging and named entity recognition (NER). For POS tagging, we use the Universal Dependencies (UD) 2.7 dataset \cite{ud-v2.7} and for NER, we use WikiANN \citep{pan-etal-2017-cross} with the train, dev and test splits from \citet{rahimi-etal-2019-massively}. In addition to these two tasks, we also use masked language modelling (MLM) on Wikipedia articles as an auxiliary task. We limit the number of sentences from Wikipedia to 100K for each language, in order to control the impact of dataset size and to reduce the training time.

For the language selection, we consider: (i) typological diversity based on language family, script and morphosyntactic attributes; (ii) a combination of high-resource and low-resource languages based on available data in downstream task; (iii) presence in the pre-training data of mBERT; and (iv) presence of a language in the two task-specific datasets.\footnote{(i) and (ii) are necessary for a realistic setting and to evaluate full-scale cross-lingual capabilities; (iii) allows us to measure if models are able to extend the limits of the MMT; (iv) enables us to assess supervision from a mixture of task and language combinations.} We provide the details of the language and dataset selection in Appendix~\ref{app:languages}.

\paragraph{Experimental Setup} We evaluate Hyper-X for zero-shot transfer in three different settings: \textbf{(1) English single-task,} where we train the models only on English data for each downstream task separately. \textbf{(2) English multi-task,} where the models are trained on English POS and NER data at the same time. \textbf{(3) Mixed-language multi-task,} where we train the models in a multi-task setup, but instead of using only English data for both POS and NER, we use a mixture of task-language combinations. In order to measure zero-shot performance in this setup, following \citet{ponti-etal-2021-parameter} we create two different partitions from all possible language-task combinations in such a way that a task-language pair is always unseen for one of the partitions (e.g.\ NER-Turkish and POS-Arabic in Figure~\ref{fig:exp-setup}). Details of partitions and our partitioning strategy are given in Appendix \ref{app:languages}.


\subsection{Baselines and Model Variants}

\noindent \textbf{mBERT {\normalfont\citep{devlin2018bert}}} is a MMT that is pre-trained for 104 languages. We use mBERT by fine-tuning all the model parameters on the available sources. As this standard approach enables cross-lingual transfer from both a single source or a set of language-task combinations, we compare it to Hyper-X in all three settings. Moreover, we use mBERT as the base model for both Hyper-X and the other baselines.  \\

\noindent \textbf{MAD-X {\normalfont\citep{pfeiffer2020mad}}} is an adapter-based modular framework for cross-lingual transfer learning based on MMTs. It combines a task-specific adapter with language-specific adapters that are independently trained for each language using MLM. 
We train MAD-X language adapters on the same Wikipedia data that is used for Hyper-X, for all languages with a default architecture.\footnote{MAD-X also introduce `invertible adapters' that adapt token embeddings. We did not use them for simpler experimental setup. Note that, as our hypernetwork is able to generate parameters for any component, it is possible to generate invertible adapters as in MAD-X.} 
Finally, for the mixed-language setup, as the original MAD-X does not allow standard multi-task training, we train the task adapters by using multiple source languages but for NER and POS separately. We call this model \texttt{MAD-X MS}. \\

\begin{table*}[!h]
\small
\centering
\renewcommand{\arraystretch}{1.1}
\begin{tabular}{llc@{\hskip 0.3in}ccccccccc}
\toprule
&&& \multicolumn{3}{c}{Named-Entity} && \multicolumn{3}{c}{Part-of-Speech} \\
&&\textbf{\#Params /}& \multicolumn{3}{c}{Recognition} && \multicolumn{3}{c}{Tagging} \\
\noalign{\smallskip}
\cline{4-6} \cline{8-10} 
\noalign{\smallskip}
\textbf{Source} & \textbf{Method} & \textbf{Time} & \textsc{seen} & \textsc{unseen} & \textsc{all} & &  \textsc{seen} &\textsc{unseen} & \textsc{all}  \\ \midrule
&\texttt{mBERT} & 177m~/~2h & 53.4 & 40.3 & 47.3 && 66.3 & 48.9 & 58.1 \\
 &\texttt{MAD-X} & 76m~/~116h & 54.3 & \textbf{51.1} & \textbf{52.8} && 67.7 & \textbf{62.6} & \textbf{65.4} \\
English&\texttt{Hyper-X Small} & 13m~/~16h & 54.2 & 47.7 & 51.2 && 66.5 & 57.9 & 62.5 \\
(Single-Task)&\texttt{Hyper-X Base} & 76m~/~18h & \textbf{54.4} & 50.7 & 52.7 && \textbf{67.8} & 58.7 & 63.5 \\
\midrule
&\texttt{mBERT} & 177m~/~2h & 53.8 & 40.4 & 47.6 && 65.8 & 47.7 & 57.3 \\
English&\texttt{Hyper-X Small} & 13m~/~16h & 52.2 & 49.3 & 50.8 && 65.1 & 57.9 & 61.7 \\
(Multi-Task)&\texttt{Hyper-X Base} & 76m~/~18h & \textbf{54.4} & \textbf{51.1} & \textbf{52.9} && \textbf{67.0} & \textbf{59.7} & \textbf{63.6} \\
\midrule 
&\texttt{mBERT} & 177m~/~2h & 56.4 & 48.7 & 52.8 && 67.2 & 54.7 & 61.4 \\
&\texttt{PSF} & 185m~/~4h & 58.1 & 54.1 & 56.2 && 70.4 & 53.8 & 62.7 \\
& \texttt{MAD-X MS} & 76m~/~116h & 62.4 & \textbf{62.2} & \textbf{62.3} && 70.7 & \textbf{67.0} & \textbf{69.0} \\
Mixed-Language&\texttt{Hyper-X Small} & 13m~/~16h & 62.0 & 58.3 & 60.3 && 70.7 & 63.2 & 67.2 \\
(Multi-Task)&\texttt{Hyper-X Base} & 76m~/~18h & \textbf{63.3} & 61.0 & \textbf{62.3} && \textbf{71.5} & 63.8 & 67.9 \\ \bottomrule
\end{tabular}
\caption{Zero-shot cross-lingual transfer results averaged over 3 runs on Named-Entity Recognition (NER; F1) and Part-of-Speech Tagging (POS; Accuracy) for mBERT, MAD-X \cite{pfeiffer2020mad}, parameter space factorization \cite[PSF;][]{ponti-etal-2021-parameter} and Hyper-X. We highlight the best results per-setting in bold. We also report the total number of parameters and fine-tuning time for all models. Note that Hyper-X corresponds to a single model trained for each partition while MAD-X consists of $N$ independently trained adapters for each task and language. MAD-X MS refers to an adapted version of the original model trained on multiple source languages but each task separately.}
\label{tab:main}
\end{table*}

\noindent \textbf{Parameter Space Factorization {\normalfont\citep{ponti-etal-2021-parameter}}} is a Bayesian framework that learns a parameter generator from multiple tasks and languages for the softmax layer on top of a MMT. However, if a language lacks annotated training data, this model cannot learn the required latent variable for the corresponding language. Therefore, we evaluate this baseline only for the mixed-language multi-task setting using the same partitions as Hyper-X. We use the original implementation with default hyper-parameters and low-rank factorization. \\

\noindent \textbf{Model Variants} We evaluated two variants of Hyper-X in order to see the impact of Hypernetwork size: Hyper-X Base model fine-tunes 76m parameters ($d_s=192$), compatible with MAD-X in terms of total number of trainable parameters, and Hyper-X Small updates only 13m parameters ($d_s=32$). Table~\ref{tab:efficiency} shows the parameter counts together with the corresponding runtime.   

\subsection{Training Details}
For all the experiments, we used a batch size of 32 and a maximum sequence length of 256. We trained Hyper-X for 100,000 updates steps by using a linearly decreasing learning rate of 1e-4 with 4000 warm-up steps. We evaluated checkpoints every 5,000 steps, and used the best checkpoint w.r.t. the average validation score for testing. As for baselines, we trained mBERT and MAD-X tasks adapters for 20 epochs by using learning rate of 1e-5 and 1e-4 respectively with the same scheduler and warm-up steps. Since MAD-X requires prerequisite language adapters, we trained language adapters for 100,000 steps for each language separately.  

In terms of model size, we use a bottleneck dimension of 256 to learn adapters for Hyper-X. Similarly, we train language and adapters with dimension of 256 and 48 for MAD-X to create a comparable baseline. In Hyper-X, as input to the hypernetwork, dimensions for task, language and layer embeddings are all set to 64 (total 192). 
During training, we create homogeneous mini-batches for each task-language combination to learn the corresponding embeddings together with the hypernetwork. Moreover, following \citet{karimi-mahabadi-etal-2021-parameter}, we also update the original layer-norm parameters. During multi-task training, we use temperature-based sampling with $T=5$ to balance each task-language pair during training (See Appendix \S~\ref{sec:sampling} for details).

\section{Zero-shot Transfer Results}

Table~\ref{tab:main} shows the aggregate zero-shot results in NER and POS tagging respectively. In addition to the average scores across all 15 zero-shot languages, we show the average of the 8 `seen' and 7 `unseen' languages separately with respect to language coverage of mBERT. We present results for English single-task, English multi-task and Mixed-language multi-task settings. 

Overall, Hyper-X Base performs on par with the strongest baseline when transferring from English. In the presence of additional sources, such as a mixture of task-language pairs, Hyper-X outperforms both mBERT and parameter space factorization (PSF). 
%
In comparison to MAD-X, Hyper-X generally performs better on seen languages. We relate this to the unified hypernetwork enabling maximum sharing between languages and higher utilization of the pre-trained capacity in contrast to the isolated adapters. On unseen languages, Hyper-X is outperformed by MAD-X in most cases. However, we emphasize that MAD-X requires training separate language adapters for each new language, which makes it considerably less resource-efficient than Hyper-X (see \S~\ref{sec:efficiency}). 

\paragraph{English Single-Task} When English is used as the only source language for each task separately, Hyper-X (Base) performs on par with MAD-X for NER (52.7 vs 52.8 F1) but falls behind for POS tagging (63.5 vs 65.4 Acc.)\ on average. Both models significantly outperform mBERT. Looking at the individual language results, Hyper-X performs slightly better on `seen' languages compared to MAD-X in NER and POS tagging respectively. 
For `unseen' languages, both MAD-X and Hyper-X benefit from MLM, which results in large improvements with respect to mBERT. Between the two models, MAD-X achieves a higher average score in both NER and POS tagging.

\paragraph{English Multi-Task} In a multi-task setting where only English data is available, fine-tuning mBERT for both target tasks at the same time gives mixed results compared to single-task training---in line with previous findings noting catastrophic forgetting and interference in MMTs \citep{wang-etal-2020-negative}. Hyper-X Base, on the other hand, shows a small but consistent improvement on the majority of languages, with 0.2 (F1) and 0.1 (Acc.) average increase in NER and POS tagging respectively. This confirms that Hyper-X is able to mitigate interference while allowing for sharing between tasks when enough capacity is provided.\footnote{MAD-X learns independent adapters for each target task, which does not allow for positive cross-task transfer.} 

\paragraph{Mixed-Language Multi-Task} In this setting, a mixture of language data is provided for NER and POS via two separate training partitions while keeping each task-language pair unseen in one of these partitions. 
All the models including mBERT achieve better zero-shot scores compared to the previous settings. Among the baselines, parameter space factorization (PSF) gives a larger improvement compared to mBERT on both tasks, indicating the importance of task- and language-specific parametrization for adapting a MMT.
Hyper-X Base produces the largest performance gain among the models that trains only a single model: it achieves 9.0 (F1) and 4.3 (Acc.) average increase for NER and POS. Although both PSF and Hyper-X enable adaptation conditioned on a mixture of task and language combinations, we relate the difference between PSF and Hyper-X to the contrast in parameter generation. PSF only generates parameters of the softmax layer and is thus unable to adapt deeper layers of the model. Hyper-X, on the other hand, generates adapter layer parameters inserted throughout the model, which provide a higher degree of adaptation flexibility. Hyper-X outperforms PSF particularly on unseen languages as it benefits from MLM as an auxiliary task.

Finally, Hyper-X tends to perform slightly better on seen languages compared to the adapted multi-source version of MAD-X. However, MAD-X outperforms Hyper-X on unseen languages by 1.2 (F1) and 2.8 (Acc.) for NER and POS respectively.
Besides the expected benefits of independently trained language adapters in MAD-X, we relate this to the limited cross-task supervision for unseen languages in Hyper-X for this setting. Especially, when the target task is POS, most of the unseen languages have only 100 sentences available in NER dataset, which leaves only a little margin for improvements. 

\begin{table}[t]
\small
\begin{tabular}{@{}lcc@{}}
\toprule
Model & \#Params.  & Training Time \\ \midrule
mBERT   & 177m & 2h  \\
PSF   & 185m & 4h  \\
\noalign{\smallskip} 
\hdashline 
\noalign{\smallskip}
MAD-X &  76m & 116h  \\ 
\noalign{\smallskip}
$\hookrightarrow$ Language Adapters & 4.7m x $l$ & 7h x $l$ \\
$\hookrightarrow$ Task Adapters & 0.9m x $t$ & 2h x $t$ \\
\noalign{\smallskip} 
\hdashline 
\noalign{\smallskip}
Hyper-X Small &  13m & 16h \\ 
Hyper-X Base &  76m & 18h \\ 
\bottomrule
\end{tabular}
\caption{Compute efficiency with respect to number of fine-tuned parameters and training time for mBERT, PSF, MAD-X and Hyper-X. Training time includes both NER and POS-tagging. For MAD-X, the total number of parameters and training time is calculated for 16 ($l$) languages and 2 ($t$) tasks.}
\label{tab:efficiency}
\end{table}

\section{Analysis}


\subsection{Parameter and Time Efficiency}
\label{sec:efficiency}

Table~\ref{tab:efficiency} shows the fine-tuned parameter counts and the training time required for the baselines and Hyper-X models. Unlike mBERT, PSF and Hyper-X, MAD-X consists of 16 and 2 independently trained language and task adapters respectively. 
In terms of parameter efficiency, MAD-X and Hyper-X Base models correspond to 43\% of mBERT's parameters. 
However, in terms of training time, Hyper-X Base is trained only once for about 18 hours, as opposed to MAD-X's considerably high total training time (116 hours in total). Thus, considering the competitive zero-shot performances across different languages and settings, Hyper-X Base provides a better efficiency-performance trade-off.
Furthermore, in the case of adding more languages, MAD-X's parameter count and training time increase linearly with the number of new languages, while Hyper-X's computational cost remains the same. 

As Hyper-X model variants, we evaluated two different sizes of the source embedding ($d_s$;~32$\rightarrow$192). 
Although Hyper-X Small is much more parameter-efficient (7.2\% of mBERT's parameters) and takes slightly less time to train (16h), its zero-shot performance is significantly lower than the base model, especially for unseen languages. Nevertheless, Hyper-X Small remains a valid alternative for particularly `seen' languages. 

\begin{figure}[t]
    \vspace{-0.4cm}
    \centering \hspace{-0.4cm}
    \includegraphics[scale=0.195]{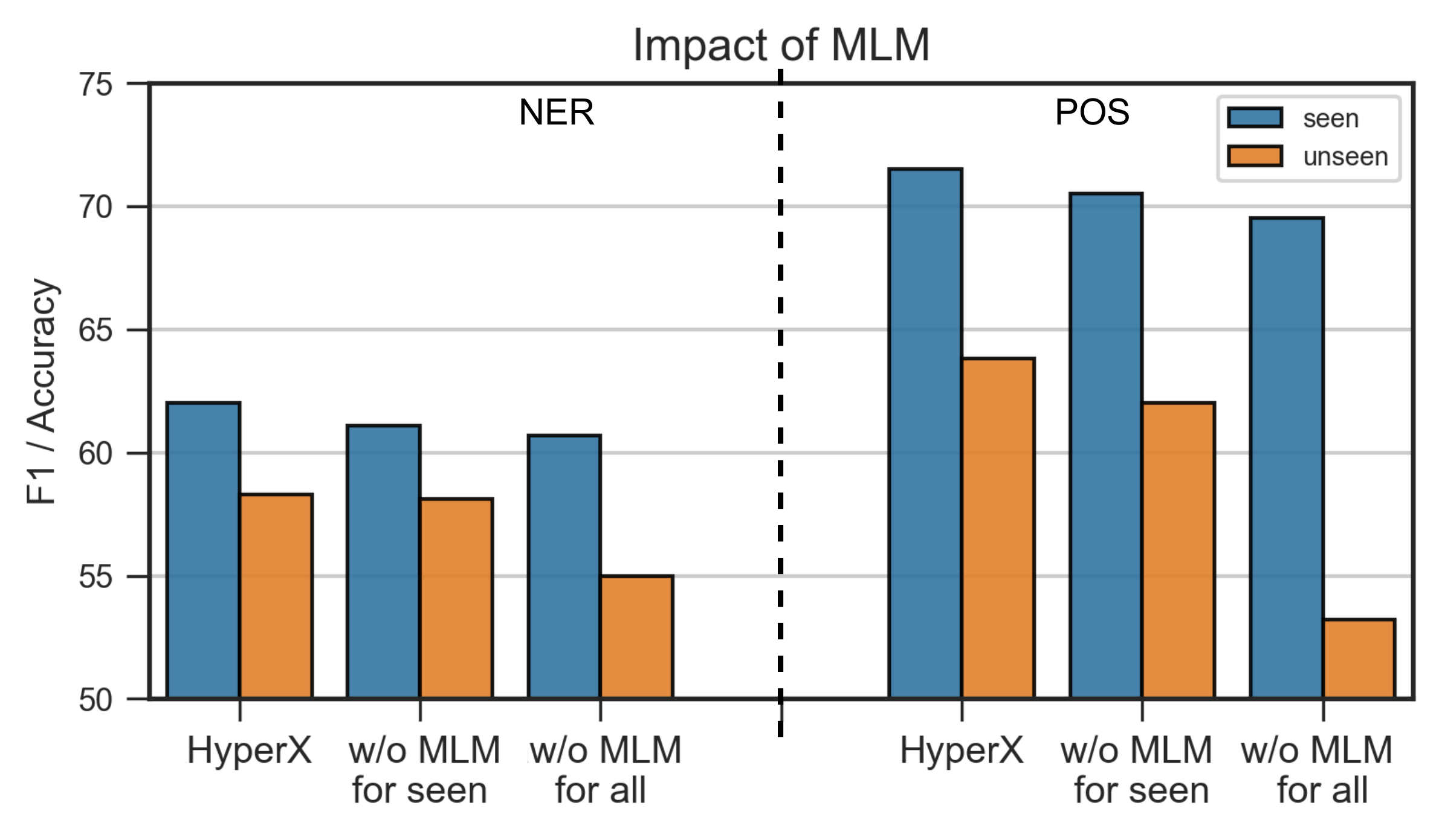}
    \caption{Impact of auxiliary MLM traning on zero-shot results for \textsc{seen} and \textsc{unseen} language groups on NER and POS tagging, when MLM data removed from the corresponding groups incrementally.}
  \label{fig:mlm}
\end{figure}

\subsection{Impact of Auxiliary MLM Training}
\label{sec:mlm}

Figure~\ref{fig:mlm} demonstrates the impact of auxiliary MLM training in Hyper-X Base for the mixed-language multi-task setting. As this setting provides training instances for each task and language, we evaluated the impact of MLM by removing the corresponding Wikipedia data first for `seen' languages, then for `all' languages. As shown in the figure, although the availability of MLM data slightly increases seen language performance, it mainly boosts the scores in unseen languages: +6.2 F1 and +10.5 Acc. for NER and POS respectively. Furthermore, when MLM data is removed for only seen languages, Hyper-X can mostly recover performance on seen languages, confirming the dominant effect of MLM on unseen languages. 

\begin{figure}[t]
    \vspace{-0.4cm}
    \centering \hspace{-0.4cm}
    \includegraphics[scale=0.195]{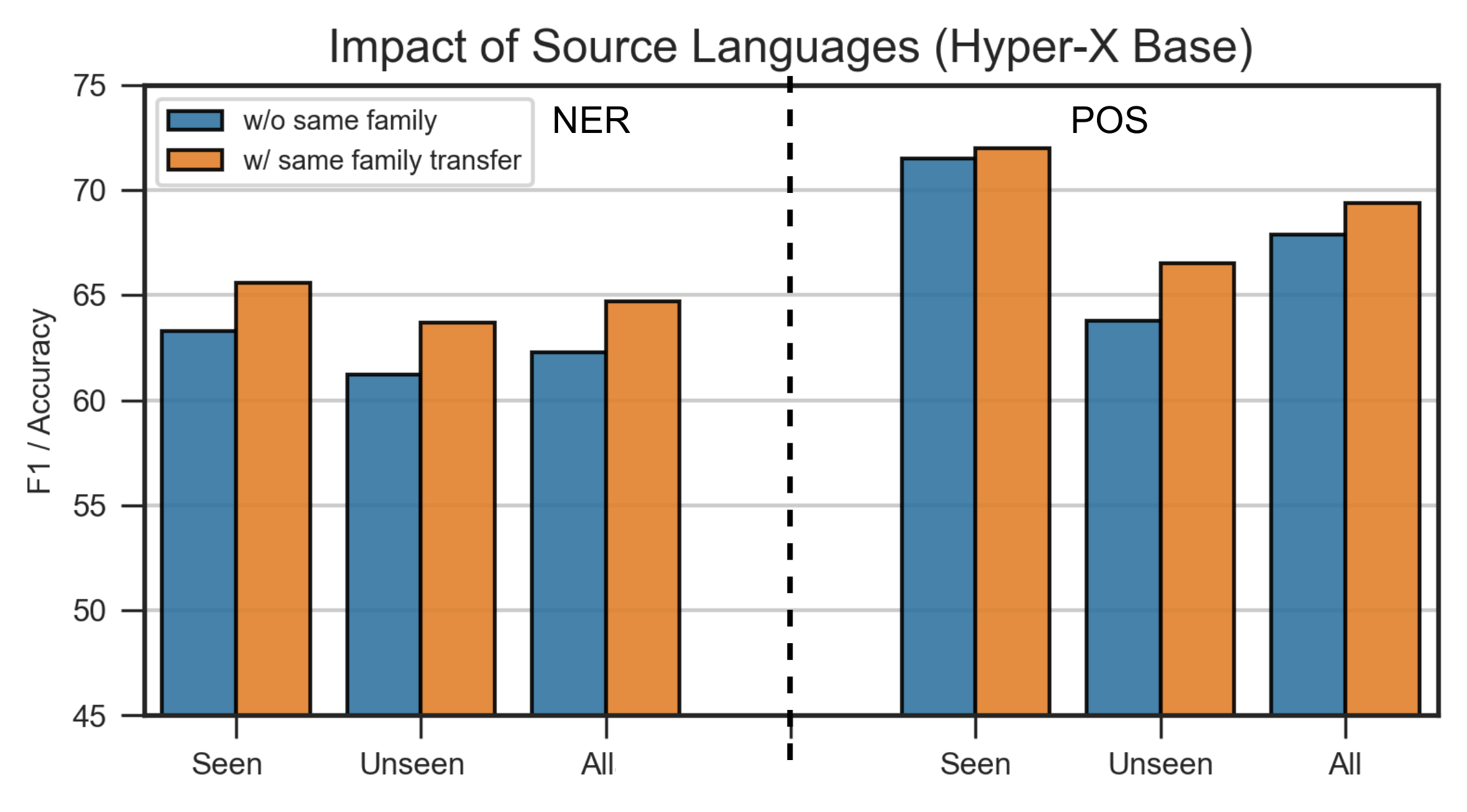}
    \caption{Impact of source language for Hyper-X Base performance on \textsc{seen}, \textsc{unseen} language groups in mixed-language multi-task setup.}
  \label{fig:hyperx-family}
\end{figure}

\subsection{Impact of Source Languages}
\label{sec:source-langs}

\begin{figure}[t]
    \vspace{-0.4cm}
    \centering \hspace{-0.4cm}
    \includegraphics[scale=0.195]{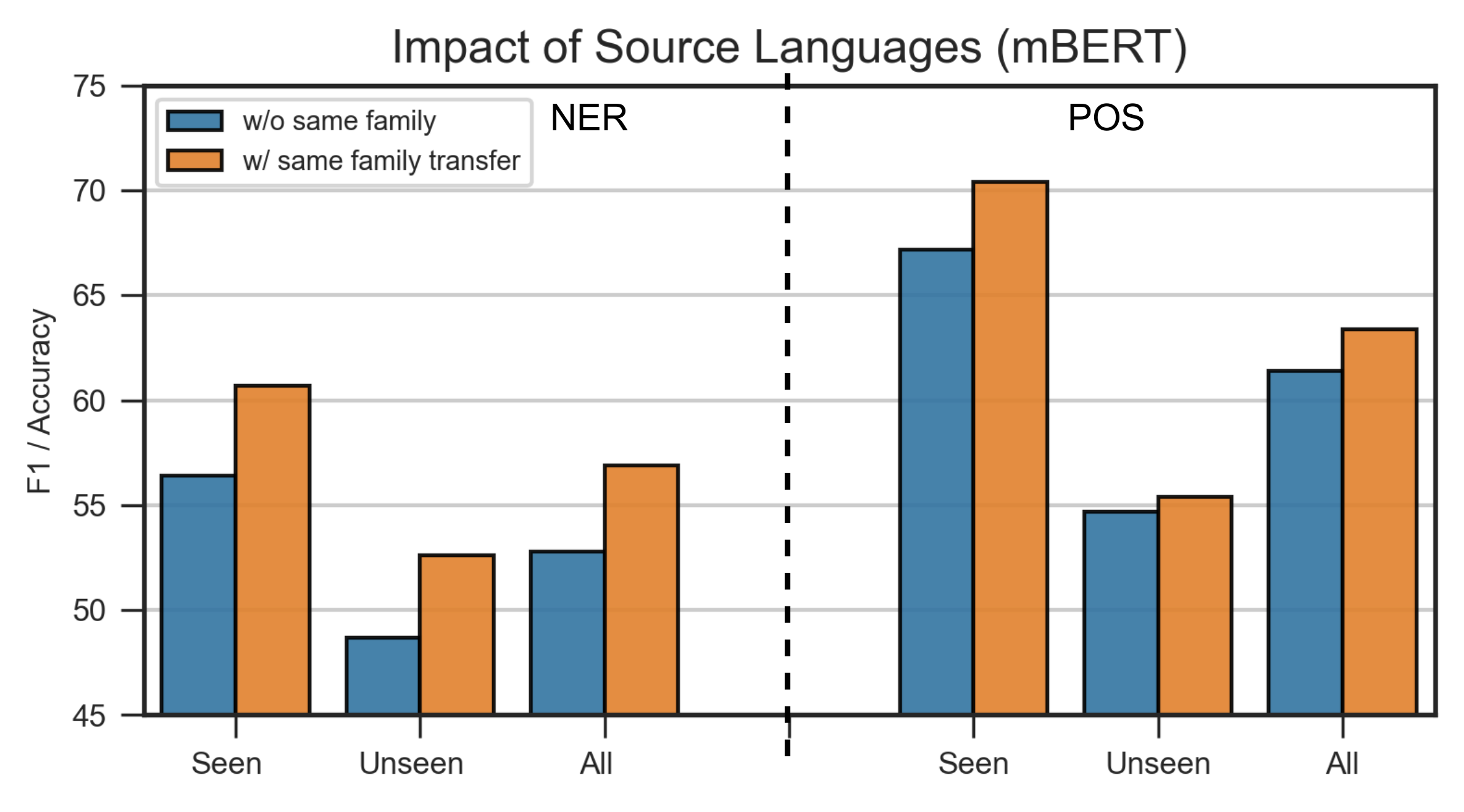}
    \caption{Impact of source language for mBERT performance on \textsc{seen}, \textsc{unseen} language groups in mixed-language multi-task setup.}
  \label{fig:mbert-family}
\end{figure}

In the mixed-language multi-task setting, we deliberately avoid grouping languages from same families to different partitions, in order to restrict the transfer from the same-language family instances, and to observe the effect of cross-task supervision. However, we also evaluate the impact of source languages in this setup, to measure the degree of potential positive transfer. To this end, we switched the partitions of \texttt{kk,mt,yue}, so that all of them will likely benefit from a high-resource language from the same family for the same target task. Figure~\ref{fig:hyperx-family} and \ref{fig:mbert-family} shows the aggregated results in both Hyper-X Base and mBERT. Firstly, both models benefit from positive transfer. Secondly, although the relative increase in mBERT is slightly higher Hyper-X still outperforms mBERT with a large margin, showing the robustness of our model with regard to different partitions. 

\subsection{Few-shot Transfer}

Fine-tuning an MMT with a few target instances has been shown to increase zero-shot performances \citep{lauscher-etal-2020-zero}. Therefore, we evaluate Hyper-X for few-shot transfer on 5 languages---3 of which are high-resource and covered by mBERT and 2 are low-resource and unseen. To this end, we further fine-tune Hyper-X and the corresponding baselines that are trained initially in the English multi-task by using 5, 10, 20, and 50 training instances for each language separately on NER and POS-tagging (see details in Appendix \S \ref{sec:few-shot-exp}).  

Figure~\ref{fig:few-shot} presents the average results comparing mBERT to MAD-X. 
Similar to the zero-shot results, on seen languages, Hyper-X constantly provides better adaptation than both baselines for NER and POS. On unseen languages, MAD-X gives the best result on average. This is because MAD-X starts with better initial representations for Maltese and Uyghur. When more samples are provided Hyper-X reduces the initial gap.
Overall, Hyper-X consistently achieves the best or competitive performance on the majority of the experiments, except `unseen' languages for POS tagging, showing the effectiveness of our approach beyond the standard zero-shot transfer. Taken together with the parameter and training efficiency, these results show that Hyper-X can be easily extended to new languages without incurring large computing costs. 

\begin{figure*}[t]
  \begin{subfigure}[b]{0.245\textwidth}
    \includegraphics[width=\textwidth]{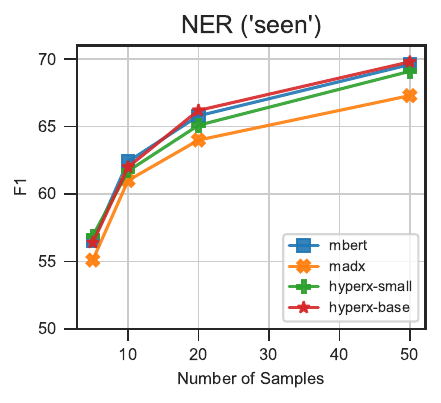}
  \end{subfigure} 
  \begin{subfigure}[b]{0.245\textwidth}
    \includegraphics[width=\textwidth]{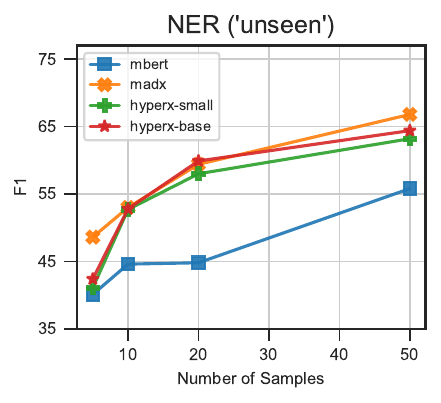}
  \end{subfigure} 
  \begin{subfigure}[b]{0.245\textwidth}
    \includegraphics[width=\textwidth]{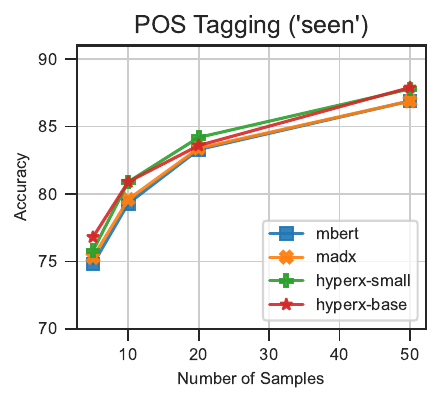}
  \end{subfigure} 
  \begin{subfigure}[b]{0.245\textwidth}
    \includegraphics[width=\textwidth]{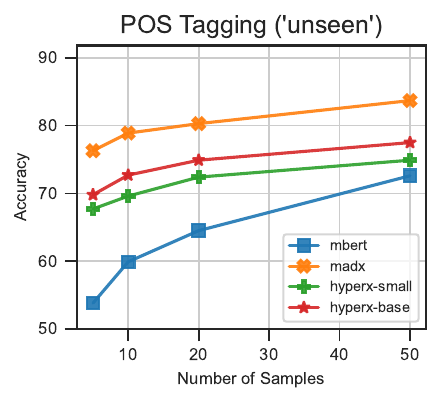}
  \end{subfigure} 
  \caption{Few-shot transfer for 5 new languages on NER, POS-tagging. Results are averaged over \textsc{seen} (\texttt{ar,tr,zh}) and \textsc{unseen} (\texttt{mt,ug}) languages. In first three settings, \textit{both} Hyper-X models competitive or better than other models. Results for all few-shot experiments are given in Appendix~\ref{sec:few-shot-exp}}
  \label{fig:few-shot}
\end{figure*}

\section{Related Work}

\paragraph{Adapters} As a parameter-efficient alternative to standard fine-tuning, adapters have been used for quick training \citep{ruckle-etal-2021-adapterdrop}, multi-task learning \citep{stickland2019bert} and knowledge composition \citep{,pfeiffer-etal-2021-adapterfusion, wang-etal-2021-k,poth-etal-2021-pre}. Moreover, \citet{mahabadi2021compacter} and \citet{he2021towards} extended adapters for better performance with fewer parameters.
In the context of multilingual transfer, adapters enable allocation of additional language-specific capacity, thereby mitigating the `curse of multilinguality' \citep{ustun-etal-2020-udapter}. Such language adapters \citep{pfeiffer2020mad,ansell-etal-2021-mad-g} achieve high zero-shot results when combined with task adapters and enable generalization to languages unseen during pre-training via MLM-based adaptation \cite{pfeiffer-etal-2021-unks}. \citet{philip-etal-2020-monolingual} and \citet{ustun-etal-2021-multilingual} also used monolingual adapters for zero-shot and unsupervised NMT. 

\paragraph{Hypernetworks in NLP} \citet{Tay2021hypergrid} propose a multi-task model that uses a hypernetwork to condition on input to learn task-specific reparametrizations. Similarly, \citet{karimi-mahabadi-etal-2021-parameter} generate task-specific adapters via a hypernetwork. Recently, \citet{He2022hyperprompt} use a hypernetwork to generate prompts. For multilingual learning, where the input sources 
correspond to language embeddings, \citet{ustun-etal-2020-udapter} and \citet{ansell-etal-2021-mad-g} learn these embeddings from the typological feature vectors of languages, enabling generalization to unseen languages based on a hypernetwork. In a similar spirit to our work,
parameter space factorization \cite[PSF;][]{ponti-etal-2021-parameter}, learns task and language-specific embeddings from seen task-language combinations. However, unlike our model, these embeddings are used for task/language-specific parametrization in the softmax layer. 



\section{Conclusion}

We have proposed Hyper-X, a novel approach for multi-task multilingual transfer learning, based on a unified hypernetwork that leverages heterogeneous sources of information, such as multiple tasks and languages.
By learning to generate composite adapters for each task-language combinations that modify the parameters of a pre-trained multilingual transformer,
Hyper-X allows for maximum information sharing and enables zero-shot prediction for arbitrary task-language pairs at test time.
Through a number of experiments, we demonstrate that Hyper-X is competitive with the state-of-the-art when transferring from a source language.
When a mixture of tasks and languages is available, Hyper-X outperforms several strong baselines on many languages,
while being more parameter and time efficient. Finally, we show that for few-shot transfer, Hyper-X is a strong option with a less computing cost than baselines for the initial task adaptation.

\section{Limitations} 



Firstly, although our experiments show the potential of Hyper-X to benefit from multiple tasks for zero-shot transfer, so far we evaluated our model on a limited set of tasks: NER and POS-tagging, which may limit the generalizability of our model to other tasks.

Secondly, for the few-shot transfer, we limit our experiments to languages that we learn via MLM and to existing tasks. Our work does not include languages without MLM data as well as completely new tasks. Learning the task and language embeddings separately, however, creates a possibility to interpolate existing embeddings for new languages or new tasks, which  especially may work for the few-shot learning. We leave exploration of these two limitations to future work. 

\section*{Acknowledgements}
We would like to thank Noah Constant, Asa Cooper Stickland and the anonymous reviewers for their helpful feedback on a previous version of this paper. We also would like to thank the Center for Information Technology of the University of Groningen for providing access to the Peregrine HPC cluster.

\bibliography{anthology,custom}

\begin{thebibliography}{38}
\expandafter\ifx\csname natexlab\endcsname\relax\def\natexlab#1{#1}\fi

\bibitem[{Ansell et~al.(2021)Ansell, Ponti, Pfeiffer, Ruder, Glava{\v{s}},
  Vuli{\'c}, and Korhonen}]{ansell-etal-2021-mad-g}
Alan Ansell, Edoardo~Maria Ponti, Jonas Pfeiffer, Sebastian Ruder, Goran
  Glava{\v{s}}, Ivan Vuli{\'c}, and Anna Korhonen. 2021.
\newblock \href {https://doi.org/10.18653/v1/2021.findings-emnlp.410}
  {{MAD}-{G}: {M}ultilingual adapter generation for efficient cross-lingual
  transfer}.
\newblock In \emph{Findings of the Association for Computational Linguistics:
  EMNLP 2021}, pages 4762--4781, Punta Cana, Dominican Republic. Association
  for Computational Linguistics.

\bibitem[{Bapna and Firat(2019)}]{bapna-firat-2019-simple}
Ankur Bapna and Orhan Firat. 2019.
\newblock \href {https://doi.org/10.18653/v1/D19-1165} {Simple, scalable
  adaptation for neural machine translation}.
\newblock In \emph{Proceedings of the 2019 Conference on Empirical Methods in
  Natural Language Processing and the 9th International Joint Conference on
  Natural Language Processing (EMNLP-IJCNLP)}, pages 1538--1548, Hong Kong,
  China. Association for Computational Linguistics.

\bibitem[{Conneau et~al.(2020)Conneau, Khandelwal, Goyal, Chaudhary, Wenzek,
  Guzm{\'a}n, Grave, Ott, Zettlemoyer, and Stoyanov}]{conneau2019unsupervised}
Alexis Conneau, Kartikay Khandelwal, Naman Goyal, Vishrav Chaudhary, Guillaume
  Wenzek, Francisco Guzm{\'a}n, Edouard Grave, Myle Ott, Luke Zettlemoyer, and
  Veselin Stoyanov. 2020.
\newblock \href {https://doi.org/10.18653/v1/2020.acl-main.747} {Unsupervised
  cross-lingual representation learning at scale}.
\newblock In \emph{Proceedings of the 58th Annual Meeting of the Association
  for Computational Linguistics}, pages 8440--8451.

\bibitem[{Devlin et~al.(2019)Devlin, Chang, Lee, and
  Toutanova}]{devlin2018bert}
Jacob Devlin, Ming-Wei Chang, Kenton Lee, and Kristina Toutanova. 2019.
\newblock \href {https://doi.org/10.18653/v1/N19-1423} {{BERT}: Pre-training of
  deep bidirectional transformers for language understanding}.
\newblock In \emph{Proceedings of the 2019 Conference of the North {A}merican
  Chapter of the Association for Computational Linguistics: Human Language
  Technologies, Volume 1 (Long and Short Papers)}, pages 4171--4186.

\bibitem[{Ha et~al.(2016)Ha, Dai, and Le}]{ha2016hypernetworks}
David Ha, Andrew Dai, and Quoc~V Le. 2016.
\newblock \href {arXiv preprint arXiv:1609.09106} {Hypernetworks}.
\newblock In \emph{International Conference on Learning Representations}.

\bibitem[{He et~al.(2022{\natexlab{a}})He, Zhou, Ma, Berg-Kirkpatrick, and
  Neubig}]{he2021towards}
Junxian He, Chunting Zhou, Xuezhe Ma, Taylor Berg-Kirkpatrick, and Graham
  Neubig. 2022{\natexlab{a}}.
\newblock Towards a unified view of parameter-efficient transfer learning.
\newblock In \emph{Proceedings of ICLR 2022}.

\bibitem[{He et~al.(2022{\natexlab{b}})He, Zheng, Tay, Gupta, Du, Aribandi,
  Zhao, Li, Chen, Metzler, Cheng, and Chi}]{He2022hyperprompt}
Yun He, Huaixiu~Steven Zheng, Yi~Tay, Jai Gupta, Yu~Du, Vamsi Aribandi, Zhe
  Zhao, YaGuang Li, Zhao Chen, Donald Metzler, Heng-Tze Cheng, and Ed~H. Chi.
  2022{\natexlab{b}}.
\newblock \href {http://arxiv.org/abs/2203.00759} {{HyperPrompt: Prompt-based
  Task-Conditioning of Transformers}}.
\newblock \emph{arXiv preprint arXiv:2203.00759}.

\bibitem[{Houlsby et~al.(2019)Houlsby, Giurgiu, Jastrzebski, Morrone,
  De~Laroussilhe, Gesmundo, Attariyan, and Gelly}]{houlsby2019parameter}
Neil Houlsby, Andrei Giurgiu, Stanislaw Jastrzebski, Bruna Morrone, Quentin
  De~Laroussilhe, Andrea Gesmundo, Mona Attariyan, and Sylvain Gelly. 2019.
\newblock \href {https://openreview.net/forum?id=H1NjknZO-H}
  {Parameter-efficient transfer learning for nlp}.
\newblock In \emph{International Conference on Machine Learning}, pages
  2790--2799.

\bibitem[{Hu et~al.(2020)Hu, Ruder, Siddhant, Neubig, Firat, and
  Johnson}]{Hu2020xtreme}
Junjie Hu, Sebastian Ruder, Aditya Siddhant, Graham Neubig, Orhan Firat, and
  Melvin Johnson. 2020.
\newblock \href {http://arxiv.org/abs/arXiv:2003.11080v1} {{XTREME: A Massively
  Multilingual Multi-task Benchmark for Evaluating Cross-lingual
  Generalization}}.
\newblock In \emph{Proceedings of ICML 2020}.

\bibitem[{Lauscher et~al.(2020{\natexlab{a}})Lauscher, Ravishankar,
  Vuli{\'{c}}, and Glava{\v{s}}}]{Lauscher2020}
Anne Lauscher, Vinit Ravishankar, Ivan Vuli{\'{c}}, and Goran Glava{\v{s}}.
  2020{\natexlab{a}}.
\newblock \href {http://arxiv.org/abs/2005.00633} {{From Zero to Hero: On the
  Limitations of Zero-Shot Cross-Lingual Transfer with Multilingual
  Transformers}}.
\newblock In \emph{Proceedings of EMNLP 2020}.

\bibitem[{Lauscher et~al.(2020{\natexlab{b}})Lauscher, Ravishankar, Vuli{\'c},
  and Glava{\v{s}}}]{lauscher-etal-2020-zero}
Anne Lauscher, Vinit Ravishankar, Ivan Vuli{\'c}, and Goran Glava{\v{s}}.
  2020{\natexlab{b}}.
\newblock \href {https://doi.org/10.18653/v1/2020.emnlp-main.363} {From zero to
  hero: {O}n the limitations of zero-shot language transfer with multilingual
  {T}ransformers}.
\newblock In \emph{Proceedings of the 2020 Conference on Empirical Methods in
  Natural Language Processing (EMNLP)}, pages 4483--4499, Online. Association
  for Computational Linguistics.

\bibitem[{Mahabadi et~al.(2021{\natexlab{a}})Mahabadi, Henderson, and
  Ruder}]{mahabadi2021compacter}
Rabeeh~Karimi Mahabadi, James Henderson, and Sebastian Ruder.
  2021{\natexlab{a}}.
\newblock Compacter: Efficient low-rank hypercomplex adapter layers.
\newblock In \emph{Advances in neural information processing systems}.

\bibitem[{Mahabadi et~al.(2021{\natexlab{b}})Mahabadi, Ruder, Dehghani, and
  Henderson}]{karimi-mahabadi-etal-2021-parameter}
Rabeeh~Karimi Mahabadi, Sebastian Ruder, Mostafa Dehghani, and James Henderson.
  2021{\natexlab{b}}.
\newblock \href {https://doi.org/10.18653/v1/2021.acl-long.47}
  {Parameter-efficient multi-task fine-tuning for transformers via shared
  hypernetworks}.
\newblock In \emph{Proceedings of the 59th Annual Meeting of the Association
  for Computational Linguistics and the 11th International Joint Conference on
  Natural Language Processing (Volume 1: Long Papers)}, pages 565--576, Online.
  Association for Computational Linguistics.

\bibitem[{McCloskey and Cohen(1989)}]{mccloskey1989catastrophic}
Michael McCloskey and Neal~J Cohen. 1989.
\newblock Catastrophic interference in connectionist networks: The sequential
  learning problem.
\newblock In \emph{Psychology of learning and motivation}, volume~24, pages
  109--165. Elsevier.

\bibitem[{Nivre et~al.(2018)Nivre, Abrams, Agi{\'c}, Ahrenberg, Antonsen,
  Aplonova, Aranzabe, Arutie, Asahara, Ateyah, Attia, Atutxa, Augustinus,
  Badmaeva, Ballesteros, Banerjee, Bank, Barbu~Mititelu, Basmov, Bauer,
  Bellato, Bengoetxea, Berzak, Bhat, Bhat, Biagetti, Bick, Blokland, Bobicev,
  B{\"o}rstell, Bosco, Bouma, Bowman, Boyd, Burchardt, Candito, Caron, Caron,
  Cebiro{\u g}lu~Eryi{\u g}it, Cecchini, Celano, {\v C}{\'e}pl{\"o}, Cetin,
  Chalub, Choi, Cho, Chun, Cinkov{\'a}, Collomb, {\c C}{\"o}ltekin, Connor,
  Courtin, Davidson, de~Marneffe, de~Paiva, Diaz~de Ilarraza, Dickerson, Dirix,
  Dobrovoljc, Dozat, Droganova, Dwivedi, Eli, Elkahky, Ephrem, Erjavec,
  Etienne, Farkas, Fernandez~Alcalde, Foster, Freitas, Gajdo{\v s}ov{\'a},
  Galbraith, Garcia, G{\"a}rdenfors, Garza, Gerdes, Ginter, Goenaga, Gojenola,
  G{\"o}k{\i}rmak, Goldberg, G{\'o}mez~Guinovart, Gonz{\'a}les~Saavedra,
  Grioni, Gr{\=u}z{\={\i}}tis, Guillaume, Guillot-Barbance, Habash, Haji{\v c},
  Haji{\v c}~jr., H{\`a}~M{\~y}, Han, Harris, Haug, Hladk{\'a}, Hlav{\'a}{\v
  c}ov{\'a}, Hociung, Hohle, Hwang, Ion, Irimia, Ishola, Jel{\'{\i}}nek,
  Johannsen, J{\o}rgensen, Ka{\c s}{\i}kara, Kahane, Kanayama, Kanerva, Katz,
  Kayadelen, Kenney, Kettnerov{\'a}, Kirchner, Kopacewicz, Kotsyba, Krek, Kwak,
  Laippala, Lambertino, Lam, Lando, Larasati, Lavrentiev, Lee,
  L{\^e}~H{\`{\^o}}ng, Lenci, Lertpradit, Leung, Li, Li, Li, Lim, Ljube{\v
  s}i{\'c}, Loginova, Lyashevskaya, Lynn, Macketanz, Makazhanov, Mandl,
  Manning, Manurung, M{\u a}r{\u a}nduc, Mare{\v c}ek, Marheinecke,
  Mart{\'{\i}}nez~Alonso, Martins, Ma{\v s}ek, Matsumoto, {McDonald}, Mendon{\c
  c}a, Miekka, Misirpashayeva, Missil{\"a}, Mititelu, Miyao, Montemagni, More,
  Moreno~Romero, Mori, Mori, Mortensen, Moskalevskyi, Muischnek, Murawaki,
  M{\"u}{\"u}risep, Nainwani, Navarro~Hor{\~n}iacek, Nedoluzhko, Ne{\v
  s}pore-B{\=e}rzkalne, Nguy{\~{\^e}}n~Th{\d i}, Nguy{\~{\^e}}n Th{\d i}~Minh,
  Nikolaev, Nitisaroj, Nurmi, Ojala, Ol{\'u}{\`o}kun, Omura, Osenova,
  {\"O}stling, {\O}vrelid, Partanen, Pascual, Passarotti, Patejuk,
  Paulino-Passos, Peng, Perez, Perrier, Petrov, Piitulainen, Pitler, Plank,
  Poibeau, Popel, Pretkalni{\c n}a, Pr{\'e}vost, Prokopidis,
  Przepi{\'o}rkowski, Puolakainen, Pyysalo, R{\"a}{\"a}bis, Rademaker,
  Ramasamy, Rama, Ramisch, Ravishankar, Real, Reddy, Rehm, Rie{\ss}ler,
  Rinaldi, Rituma, Rocha, Romanenko, Rosa, Rovati, Roșca, Rudina, Rueter,
  Sadde, Sagot, Saleh, Samard{\v z}i{\'c}, Samson, Sanguinetti, Saul{\={\i}}te,
  Sawanakunanon, Schneider, Schuster, Seddah, Seeker, Seraji, Shen, Shimada,
  Shohibussirri, Sichinava, Silveira, Simi, Simionescu, Simk{\'o}, {\v
  S}imkov{\'a}, Simov, Smith, Soares-Bastos, Spadine, Stella, Straka,
  Strnadov{\'a}, Suhr, Sulubacak, Sz{\'a}nt{\'o}, Taji, Takahashi, Tanaka,
  Tellier, Trosterud, Trukhina, Tsarfaty, Tyers, Uematsu, Ure{\v s}ov{\'a},
  Uria, Uszkoreit, Vajjala, van Niekerk, van Noord, Varga, Villemonte de~la
  Clergerie, Vincze, Wallin, Wang, Washington, Williams, Wir{\'e}n,
  Woldemariam, Wong, Yan, Yavrumyan, Yu, {\v Z}abokrtsk{\'y}, Zeldes, Zeman,
  Zhang, and Zhu}]{11234/1-2895}
Joakim Nivre, Mitchell Abrams, {\v Z}eljko Agi{\'c}, Lars Ahrenberg, Lene
  Antonsen, Katya Aplonova, Maria~Jesus Aranzabe, Gashaw Arutie, Masayuki
  Asahara, Luma Ateyah, Mohammed Attia, Aitziber Atutxa, Liesbeth Augustinus,
  Elena Badmaeva, Miguel Ballesteros, Esha Banerjee, Sebastian Bank, Verginica
  Barbu~Mititelu, Victoria Basmov, John Bauer, Sandra Bellato, Kepa Bengoetxea,
  Yevgeni Berzak, Irshad~Ahmad Bhat, Riyaz~Ahmad Bhat, Erica Biagetti, Eckhard
  Bick, Rogier Blokland, Victoria Bobicev, Carl B{\"o}rstell, Cristina Bosco,
  Gosse Bouma, Sam Bowman, Adriane Boyd, Aljoscha Burchardt, Marie Candito,
  Bernard Caron, Gauthier Caron, G{\"u}l{\c s}en Cebiro{\u g}lu~Eryi{\u g}it,
  Flavio~Massimiliano Cecchini, Giuseppe G.~A. Celano, Slavom{\'{\i}}r {\v
  C}{\'e}pl{\"o}, Savas Cetin, Fabricio Chalub, Jinho Choi, Yongseok Cho,
  Jayeol Chun, Silvie Cinkov{\'a}, Aur{\'e}lie Collomb, {\c C}a{\u g}r{\i} {\c
  C}{\"o}ltekin, Miriam Connor, Marine Courtin, Elizabeth Davidson,
  Marie-Catherine de~Marneffe, Valeria de~Paiva, Arantza Diaz~de Ilarraza,
  Carly Dickerson, Peter Dirix, Kaja Dobrovoljc, Timothy Dozat, Kira Droganova,
  Puneet Dwivedi, Marhaba Eli, Ali Elkahky, Binyam Ephrem, Toma{\v z} Erjavec,
  Aline Etienne, Rich{\'a}rd Farkas, Hector Fernandez~Alcalde, Jennifer Foster,
  Cl{\'a}udia Freitas, Katar{\'{\i}}na Gajdo{\v s}ov{\'a}, Daniel Galbraith,
  Marcos Garcia, Moa G{\"a}rdenfors, Sebastian Garza, Kim Gerdes, Filip Ginter,
  Iakes Goenaga, Koldo Gojenola, Memduh G{\"o}k{\i}rmak, Yoav Goldberg, Xavier
  G{\'o}mez~Guinovart, Berta Gonz{\'a}les~Saavedra, Matias Grioni, Normunds
  Gr{\=u}z{\={\i}}tis, Bruno Guillaume, C{\'e}line Guillot-Barbance, Nizar
  Habash, Jan Haji{\v c}, Jan Haji{\v c}~jr., Linh H{\`a}~M{\~y}, Na-Rae Han,
  Kim Harris, Dag Haug, Barbora Hladk{\'a}, Jaroslava Hlav{\'a}{\v c}ov{\'a},
  Florinel Hociung, Petter Hohle, Jena Hwang, Radu Ion, Elena Irimia, {\d
  O}l{\'a}j{\'{\i}}d{\'e} Ishola, Tom{\'a}{\v s} Jel{\'{\i}}nek, Anders
  Johannsen, Fredrik J{\o}rgensen, H{\"u}ner Ka{\c s}{\i}kara, Sylvain Kahane,
  Hiroshi Kanayama, Jenna Kanerva, Boris Katz, Tolga Kayadelen, Jessica Kenney,
  V{\'a}clava Kettnerov{\'a}, Jesse Kirchner, Kamil Kopacewicz, Natalia
  Kotsyba, Simon Krek, Sookyoung Kwak, Veronika Laippala, Lorenzo Lambertino,
  Lucia Lam, Tatiana Lando, Septina~Dian Larasati, Alexei Lavrentiev, John Lee,
  Phuong L{\^e}~H{\`{\^o}}ng, Alessandro Lenci, Saran Lertpradit, Herman Leung,
  Cheuk~Ying Li, Josie Li, Keying Li, {KyungTae} Lim, Nikola Ljube{\v s}i{\'c},
  Olga Loginova, Olga Lyashevskaya, Teresa Lynn, Vivien Macketanz, Aibek
  Makazhanov, Michael Mandl, Christopher Manning, Ruli Manurung, C{\u a}t{\u
  a}lina M{\u a}r{\u a}nduc, David Mare{\v c}ek, Katrin Marheinecke, H{\'e}ctor
  Mart{\'{\i}}nez~Alonso, Andr{\'e} Martins, Jan Ma{\v s}ek, Yuji Matsumoto,
  Ryan {McDonald}, Gustavo Mendon{\c c}a, Niko Miekka, Margarita
  Misirpashayeva, Anna Missil{\"a}, C{\u a}t{\u a}lin Mititelu, Yusuke Miyao,
  Simonetta Montemagni, Amir More, Laura Moreno~Romero, Keiko~Sophie Mori,
  Shinsuke Mori, Bjartur Mortensen, Bohdan Moskalevskyi, Kadri Muischnek, Yugo
  Murawaki, Kaili M{\"u}{\"u}risep, Pinkey Nainwani, Juan~Ignacio
  Navarro~Hor{\~n}iacek, Anna Nedoluzhko, Gunta Ne{\v s}pore-B{\=e}rzkalne,
  Luong Nguy{\~{\^e}}n~Th{\d i}, Huy{\`{\^e}}n Nguy{\~{\^e}}n Th{\d i}~Minh,
  Vitaly Nikolaev, Rattima Nitisaroj, Hanna Nurmi, Stina Ojala, Ad{\'e}day{\d
  o} Ol{\'u}{\`o}kun, Mai Omura, Petya Osenova, Robert {\"O}stling, Lilja
  {\O}vrelid, Niko Partanen, Elena Pascual, Marco Passarotti, Agnieszka
  Patejuk, Guilherme Paulino-Passos, Siyao Peng, Cenel-Augusto Perez, Guy
  Perrier, Slav Petrov, Jussi Piitulainen, Emily Pitler, Barbara Plank, Thierry
  Poibeau, Martin Popel, Lauma Pretkalni{\c n}a, Sophie Pr{\'e}vost, Prokopis
  Prokopidis, Adam Przepi{\'o}rkowski, Tiina Puolakainen, Sampo Pyysalo,
  Andriela R{\"a}{\"a}bis, Alexandre Rademaker, Loganathan Ramasamy, Taraka
  Rama, Carlos Ramisch, Vinit Ravishankar, Livy Real, Siva Reddy, Georg Rehm,
  Michael Rie{\ss}ler, Larissa Rinaldi, Laura Rituma, Luisa Rocha, Mykhailo
  Romanenko, Rudolf Rosa, Davide Rovati, Valentin Roșca, Olga Rudina, Jack
  Rueter, Shoval Sadde, Beno{\^{\i}}t Sagot, Shadi Saleh, Tanja Samard{\v
  z}i{\'c}, Stephanie Samson, Manuela Sanguinetti, Baiba Saul{\={\i}}te, Yanin
  Sawanakunanon, Nathan Schneider, Sebastian Schuster, Djam{\'e} Seddah,
  Wolfgang Seeker, Mojgan Seraji, Mo~Shen, Atsuko Shimada, Muh Shohibussirri,
  Dmitry Sichinava, Natalia Silveira, Maria Simi, Radu Simionescu, Katalin
  Simk{\'o}, M{\'a}ria {\v S}imkov{\'a}, Kiril Simov, Aaron Smith, Isabela
  Soares-Bastos, Carolyn Spadine, Antonio Stella, Milan Straka, Jana
  Strnadov{\'a}, Alane Suhr, Umut Sulubacak, Zsolt Sz{\'a}nt{\'o}, Dima Taji,
  Yuta Takahashi, Takaaki Tanaka, Isabelle Tellier, Trond Trosterud, Anna
  Trukhina, Reut Tsarfaty, Francis Tyers, Sumire Uematsu, Zde{\v n}ka Ure{\v
  s}ov{\'a}, Larraitz Uria, Hans Uszkoreit, Sowmya Vajjala, Daniel van Niekerk,
  Gertjan van Noord, Viktor Varga, Eric Villemonte de~la Clergerie, Veronika
  Vincze, Lars Wallin, Jing~Xian Wang, Jonathan~North Washington, Seyi
  Williams, Mats Wir{\'e}n, Tsegay Woldemariam, Tak-sum Wong, Chunxiao Yan,
  Marat~M. Yavrumyan, Zhuoran Yu, Zden{\v e}k {\v Z}abokrtsk{\'y}, Amir Zeldes,
  Daniel Zeman, Manying Zhang, and Hanzhi Zhu. 2018.
\newblock \href {http://hdl.handle.net/11234/1-2895} {Universal dependencies
  2.3}.
\newblock {LINDAT}/{CLARIAH}-{CZ} digital library at the Institute of Formal
  and Applied Linguistics ({{\'U}FAL}), Faculty of Mathematics and Physics,
  Charles University.

\bibitem[{Pan et~al.(2017)Pan, Zhang, May, Nothman, Knight, and
  Ji}]{pan-etal-2017-cross}
Xiaoman Pan, Boliang Zhang, Jonathan May, Joel Nothman, Kevin Knight, and Heng
  Ji. 2017.
\newblock \href {https://doi.org/10.18653/v1/P17-1178} {Cross-lingual name
  tagging and linking for 282 languages}.
\newblock In \emph{Proceedings of the 55th Annual Meeting of the Association
  for Computational Linguistics (Volume 1: Long Papers)}, pages 1946--1958,
  Vancouver, Canada. Association for Computational Linguistics.

\bibitem[{Pfeiffer et~al.(2021{\natexlab{a}})Pfeiffer, Kamath, R{\"u}ckl{\'e},
  Cho, and Gurevych}]{pfeiffer-etal-2021-adapterfusion}
Jonas Pfeiffer, Aishwarya Kamath, Andreas R{\"u}ckl{\'e}, Kyunghyun Cho, and
  Iryna Gurevych. 2021{\natexlab{a}}.
\newblock \href {https://doi.org/10.18653/v1/2021.eacl-main.39}
  {{A}dapter{F}usion: Non-destructive task composition for transfer learning}.
\newblock In \emph{Proceedings of the 16th Conference of the European Chapter
  of the Association for Computational Linguistics: Main Volume}, pages
  487--503, Online. Association for Computational Linguistics.

\bibitem[{Pfeiffer et~al.(2020{\natexlab{a}})Pfeiffer, R\"uckl\'{e}, Poth,
  Kamath, Vuli\'{c}, Ruder, Cho, and Gurevych}]{pfeiffer2020AdapterHub}
Jonas Pfeiffer, Andreas R\"uckl\'{e}, Clifton Poth, Aishwarya Kamath, Ivan
  Vuli\'{c}, Sebastian Ruder, Kyunghyun Cho, and Iryna Gurevych.
  2020{\natexlab{a}}.
\newblock \href {https://www.aclweb.org/anthology/2020.emnlp-demos.7}
  {Adapterhub: A framework for adapting transformers}.
\newblock In \emph{Proceedings of the 2020 Conference on Empirical Methods in
  Natural Language Processing (EMNLP 2020): Systems Demonstrations}, pages
  46--54, Online. Association for Computational Linguistics.

\bibitem[{Pfeiffer et~al.(2020{\natexlab{b}})Pfeiffer, Vuli{\'c}, Gurevych, and
  Ruder}]{pfeiffer2020mad}
Jonas Pfeiffer, Ivan Vuli{\'c}, Iryna Gurevych, and Sebastian Ruder.
  2020{\natexlab{b}}.
\newblock \href {https://arxiv.org/abs/2005.00052} {Mad-x: An adapter-based
  framework for multi-task cross-lingual transfer}.
\newblock In \emph{Proceedings of the 2020 Conference on Empirical Methods in
  Natural Language Processing}.

\bibitem[{Pfeiffer et~al.(2021{\natexlab{b}})Pfeiffer, Vuli{\'c}, Gurevych, and
  Ruder}]{pfeiffer-etal-2021-unks}
Jonas Pfeiffer, Ivan Vuli{\'c}, Iryna Gurevych, and Sebastian Ruder.
  2021{\natexlab{b}}.
\newblock \href {https://doi.org/10.18653/v1/2021.emnlp-main.800} {{UNK}s
  everywhere: {A}dapting multilingual language models to new scripts}.
\newblock In \emph{Proceedings of the 2021 Conference on Empirical Methods in
  Natural Language Processing}, pages 10186--10203, Online and Punta Cana,
  Dominican Republic. Association for Computational Linguistics.

\bibitem[{Philip et~al.(2020)Philip, Berard, Gall{\'e}, and
  Besacier}]{philip-etal-2020-monolingual}
Jerin Philip, Alexandre Berard, Matthias Gall{\'e}, and Laurent Besacier. 2020.
\newblock \href {https://doi.org/10.18653/v1/2020.emnlp-main.361} {Monolingual
  adapters for zero-shot neural machine translation}.
\newblock In \emph{Proceedings of the 2020 Conference on Empirical Methods in
  Natural Language Processing (EMNLP)}, pages 4465--4470, Online. Association
  for Computational Linguistics.

\bibitem[{Platanios et~al.(2018)Platanios, Sachan, Neubig, and
  Mitchell}]{platanios2018contextual}
Emmanouil~Antonios Platanios, Mrinmaya Sachan, Graham Neubig, and Tom Mitchell.
  2018.
\newblock \href {https://doi.org/10.18653/v1/D18-1039} {Contextual parameter
  generation for universal neural machine translation}.
\newblock In \emph{Proceedings of the 2018 Conference on Empirical Methods in
  Natural Language Processing}, pages 425--435.

\bibitem[{Ponti et~al.(2021)Ponti, Vuli{\'c}, Cotterell, Parovic, Reichart, and
  Korhonen}]{ponti-etal-2021-parameter}
Edoardo~M. Ponti, Ivan Vuli{\'c}, Ryan Cotterell, Marinela Parovic, Roi
  Reichart, and Anna Korhonen. 2021.
\newblock \href {https://doi.org/10.1162/tacl_a_00374} {Parameter space
  factorization for zero-shot learning across tasks and languages}.
\newblock \emph{Transactions of the Association for Computational Linguistics},
  9:410--428.

\bibitem[{Poth et~al.(2021)Poth, Pfeiffer, R{\"u}ckl{\'e}, and
  Gurevych}]{poth-etal-2021-pre}
Clifton Poth, Jonas Pfeiffer, Andreas R{\"u}ckl{\'e}, and Iryna Gurevych. 2021.
\newblock \href {https://doi.org/10.18653/v1/2021.emnlp-main.827} {{W}hat to
  pre-train on? {E}fficient intermediate task selection}.
\newblock In \emph{Proceedings of the 2021 Conference on Empirical Methods in
  Natural Language Processing}, pages 10585--10605, Online and Punta Cana,
  Dominican Republic. Association for Computational Linguistics.

\bibitem[{Rahimi et~al.(2019)Rahimi, Li, and Cohn}]{rahimi-etal-2019-massively}
Afshin Rahimi, Yuan Li, and Trevor Cohn. 2019.
\newblock \href {https://doi.org/10.18653/v1/P19-1015} {Massively multilingual
  transfer for {NER}}.
\newblock In \emph{Proceedings of the 57th Annual Meeting of the Association
  for Computational Linguistics}, pages 151--164, Florence, Italy. Association
  for Computational Linguistics.

\bibitem[{Rebuffi et~al.(2018)Rebuffi, Bilen, and
  Vedaldi}]{rebuffi2018efficient}
Sylvestre-Alvise Rebuffi, Hakan Bilen, and Andrea Vedaldi. 2018.
\newblock \href
  {http://openaccess.thecvf.com/content_cvpr_2018/html/Rebuffi_Efficient_Parametrization_of_CVPR_2018_paper.html}
  {Efficient parametrization of multi-domain deep neural networks}.
\newblock In \emph{Proceedings of the IEEE Conference on Computer Vision and
  Pattern Recognition}, pages 8119--8127.

\bibitem[{R{\"u}ckl{\'e} et~al.(2021)R{\"u}ckl{\'e}, Geigle, Glockner, Beck,
  Pfeiffer, Reimers, and Gurevych}]{ruckle-etal-2021-adapterdrop}
Andreas R{\"u}ckl{\'e}, Gregor Geigle, Max Glockner, Tilman Beck, Jonas
  Pfeiffer, Nils Reimers, and Iryna Gurevych. 2021.
\newblock \href {https://doi.org/10.18653/v1/2021.emnlp-main.626}
  {{AdapterDrop}: {O}n the efficiency of adapters in transformers}.
\newblock In \emph{Proceedings of the 2021 Conference on Empirical Methods in
  Natural Language Processing}, pages 7930--7946, Online and Punta Cana,
  Dominican Republic. Association for Computational Linguistics.

\bibitem[{Ruder et~al.(2019)Ruder, Peters, Swayamdipta, and
  Wolf}]{ruder2019transfer}
Sebastian Ruder, Matthew~E Peters, Swabha Swayamdipta, and Thomas Wolf. 2019.
\newblock Transfer learning in natural language processing.
\newblock In \emph{Proceedings of the 2019 conference of the North American
  chapter of the association for computational linguistics: Tutorials}, pages
  15--18.

\bibitem[{Stickland and Murray(2019)}]{stickland2019bert}
Asa~Cooper Stickland and Iain Murray. 2019.
\newblock \href {http://proceedings.mlr.press/v97/stickland19a.html} {Bert and
  pals: Projected attention layers for efficient adaptation in multi-task
  learning}.
\newblock In \emph{International Conference on Machine Learning}, pages
  5986--5995.

\bibitem[{Strassel and Tracey(2016)}]{strassel2016lorelei}
Stephanie Strassel and Jennifer Tracey. 2016.
\newblock Lorelei language packs: Data, tools, and resources for technology
  development in low resource languages.
\newblock In \emph{Proceedings of the Tenth International Conference on
  Language Resources and Evaluation (LREC'16)}, pages 3273--3280.

\bibitem[{Tay et~al.(2021)Tay, Zhao, Bahri, Metzler, and
  Juan}]{Tay2021hypergrid}
Yi~Tay, Zhe Zhao, Dara Bahri, Donald Metzler, and Da-Cheng Juan. 2021.
\newblock {HyperGrid Transformers: Towards A Single Model for Multiple Tasks}.
\newblock In \emph{Proceedings of ICLR 2021}.

\bibitem[{{\"U}st{\"u}n et~al.(2021){\"U}st{\"u}n, Berard, Besacier, and
  Gall{\'e}}]{ustun-etal-2021-multilingual}
Ahmet {\"U}st{\"u}n, Alexandre Berard, Laurent Besacier, and Matthias
  Gall{\'e}. 2021.
\newblock \href {https://doi.org/10.18653/v1/2021.emnlp-main.533} {Multilingual
  unsupervised neural machine translation with denoising adapters}.
\newblock In \emph{Proceedings of the 2021 Conference on Empirical Methods in
  Natural Language Processing}, pages 6650--6662, Online and Punta Cana,
  Dominican Republic. Association for Computational Linguistics.

\bibitem[{{\"U}st{\"u}n et~al.(2020){\"U}st{\"u}n, Bisazza, Bouma, and van
  Noord}]{ustun-etal-2020-udapter}
Ahmet {\"U}st{\"u}n, Arianna Bisazza, Gosse Bouma, and Gertjan van Noord. 2020.
\newblock \href {https://doi.org/10.18653/v1/2020.emnlp-main.180} {{UD}apter:
  Language adaptation for truly {U}niversal {D}ependency parsing}.
\newblock In \emph{Proceedings of the 2020 Conference on Empirical Methods in
  Natural Language Processing (EMNLP)}, pages 2302--2315, Online. Association
  for Computational Linguistics.

\bibitem[{Wang et~al.(2021)Wang, Tang, Duan, Wei, Huang, Ji, Cao, Jiang, and
  Zhou}]{wang-etal-2021-k}
Ruize Wang, Duyu Tang, Nan Duan, Zhongyu Wei, Xuanjing Huang, Jianshu Ji,
  Guihong Cao, Daxin Jiang, and Ming Zhou. 2021.
\newblock \href {https://doi.org/10.18653/v1/2021.findings-acl.121}
  {{K-Adapter}: {I}nfusing {K}nowledge into {P}re-{T}rained {M}odels with
  {A}dapters}.
\newblock In \emph{Findings of the Association for Computational Linguistics:
  ACL-IJCNLP 2021}, pages 1405--1418, Online. Association for Computational
  Linguistics.

\bibitem[{Wang et~al.(2020)Wang, Lipton, and
  Tsvetkov}]{wang-etal-2020-negative}
Zirui Wang, Zachary~C. Lipton, and Yulia Tsvetkov. 2020.
\newblock \href {https://doi.org/10.18653/v1/2020.emnlp-main.359} {On negative
  interference in multilingual models: Findings and a meta-learning treatment}.
\newblock In \emph{Proceedings of the 2020 Conference on Empirical Methods in
  Natural Language Processing (EMNLP)}, pages 4438--4450, Online. Association
  for Computational Linguistics.

\bibitem[{Wilie et~al.(2020)Wilie, Vincentio, Winata, Cahyawijaya, Li, Lim,
  Soleman, Mahendra, Fung, Bahar, and Purwarianti}]{wilie-etal-2020-indonlu}
Bryan Wilie, Karissa Vincentio, Genta~Indra Winata, Samuel Cahyawijaya,
  Xiaohong Li, Zhi~Yuan Lim, Sidik Soleman, Rahmad Mahendra, Pascale Fung,
  Syafri Bahar, and Ayu Purwarianti. 2020.
\newblock \href {https://aclanthology.org/2020.aacl-main.85} {{I}ndo{NLU}:
  Benchmark and resources for evaluating {I}ndonesian natural language
  understanding}.
\newblock In \emph{Proceedings of the 1st Conference of the Asia-Pacific
  Chapter of the Association for Computational Linguistics and the 10th
  International Joint Conference on Natural Language Processing}, pages
  843--857, Suzhou, China. Association for Computational Linguistics.

\bibitem[{Wolf et~al.(2020)Wolf, Debut, Sanh, Chaumond, Delangue, Moi, Cistac,
  Rault, Louf, Funtowicz, Davison, Shleifer, von Platen, Ma, Jernite, Plu, Xu,
  Le~Scao, Gugger, Drame, Lhoest, and Rush}]{wolf-etal-2020-transformers}
Thomas Wolf, Lysandre Debut, Victor Sanh, Julien Chaumond, Clement Delangue,
  Anthony Moi, Pierric Cistac, Tim Rault, Remi Louf, Morgan Funtowicz, Joe
  Davison, Sam Shleifer, Patrick von Platen, Clara Ma, Yacine Jernite, Julien
  Plu, Canwen Xu, Teven Le~Scao, Sylvain Gugger, Mariama Drame, Quentin Lhoest,
  and Alexander Rush. 2020.
\newblock \href {https://doi.org/10.18653/v1/2020.emnlp-demos.6} {Transformers:
  State-of-the-art natural language processing}.
\newblock In \emph{Proceedings of the 2020 Conference on Empirical Methods in
  Natural Language Processing: System Demonstrations}, pages 38--45, Online.
  Association for Computational Linguistics.

\bibitem[{Zeman et~al.(2020)Zeman, Nivre, Abrams, Ackermann, Aepli, Aghaei,
  Agi{\'c}, Ahmadi, Ahrenberg, Ajede, Aleksandravi{\v c}i{\=u}t{\.e}, Alfina,
  Antonsen, Aplonova, Aquino, Aragon, Aranzabe, Arnard{\'o}ttir, Arutie,
  Arwidarasti, Asahara, Ateyah, Atmaca, Attia, Atutxa, Augustinus, Badmaeva,
  Balasubramani, Ballesteros, Banerjee, Bank, Barbu~Mititelu, Basmov,
  Batchelor, Bauer, Bedir, Bengoetxea, Berk, Berzak, Bhat, Bhat, Biagetti,
  Bick, Bielinskien{\.e}, Bjarnad{\'o}ttir, Blokland, Bobicev, Boizou,
  Borges~V{\"o}lker, B{\"o}rstell, Bosco, Bouma, Bowman, Boyd, Brokait{\.e},
  Burchardt, Candito, Caron, Caron, Cavalcanti, Cebiro{\u g}lu~Eryi{\u g}it,
  Cecchini, Celano, {\v C}{\'e}pl{\"o}, Cetin, {\c C}etino{\u g}lu, Chalub,
  Chi, Cho, Choi, Chun, Cignarella, Cinkov{\'a}, Collomb, {\c C}{\"o}ltekin,
  Connor, Courtin, Davidson, de~Marneffe, de~Paiva, Derin, de~Souza, Diaz~de
  Ilarraza, Dickerson, Dinakaramani, Dione, Dirix, Dobrovoljc, Dozat,
  Droganova, Dwivedi, Eckhoff, Eli, Elkahky, Ephrem, Erina, Erjavec, Etienne,
  Evelyn, Facundes, Farkas, Fernanda, Fernandez~Alcalde, Foster, Freitas,
  Fujita, Gajdo{\v s}ov{\'a}, Galbraith, Garcia, G{\"a}rdenfors, Garza,
  Gerardi, Gerdes, Ginter, Goenaga, Gojenola, G{\"o}k{\i}rmak, Goldberg,
  G{\'o}mez~Guinovart, Gonz{\'a}lez~Saavedra, Grici{\=u}t{\.e}, Grioni, Grobol,
  Gr{\= u}z{\={\i}}tis, Guillaume, Guillot-Barbance, G{\"u}ng{\"o}r, Habash,
  Hafsteinsson, Haji{\v c}, Haji{\v c}~jr., H{\"a}m{\"a}l{\"a}inen,
  H{\`a}~M{\~y}, Han, Hanifmuti, Hardwick, Harris, Haug, Heinecke, Hellwig,
  Hennig, Hladk{\'a}, Hlav{\'a}{\v c}ov{\'a}, Hociung, Hohle, Huber, Hwang,
  Ikeda, Ingason, Ion, Irimia, Ishola, Jel{\'{\i}}nek, Johannsen,
  J{\'o}nsd{\'o}ttir, J{\o}rgensen, Juutinen, K, Ka{\c s}{\i}kara, Kaasen,
  Kabaeva, Kahane, Kanayama, Kanerva, Katz, Kayadelen, Kenney, Kettnerov{\'a},
  Kirchner, Klementieva, K{\"o}hn, K{\"o}ksal, Kopacewicz, Korkiakangas,
  Kotsyba, Kovalevskait{\.e}, Krek, Krishnamurthy, Kwak, Laippala, Lam,
  Lambertino, Lando, Larasati, Lavrentiev, Lee, L{\^e}~H{\`{\^o}}ng, Lenci,
  Lertpradit, Leung, Levina, Li, Li, Li, Li, Lim, Lind{\'e}n, Ljube{\v
  s}i{\'c}, Loginova, Luthfi, Luukko, Lyashevskaya, Lynn, Macketanz,
  Makazhanov, Mandl, Manning, Manurung, M{\u a}r{\u a}nduc, Mare{\v c}ek,
  Marheinecke, Mart{\'{\i}}nez~Alonso, Martins, Ma{\v s}ek, Matsuda, Matsumoto,
  {McDonald}, {McGuinness}, Mendon{\c c}a, Miekka, Mischenkova, Misirpashayeva,
  Missil{\"a}, Mititelu, Mitrofan, Miyao, Mojiri~Foroushani, Moloodi,
  Montemagni, More, Moreno~Romero, Mori, Mori, Morioka, Moro, Mortensen,
  Moskalevskyi, Muischnek, Munro, Murawaki, M{\"u}{\"u}risep, Nainwani,
  Nakhl{\'e}, Navarro~Hor{\~n}iacek, Nedoluzhko, Ne{\v s}pore-B{\=e}rzkalne,
  Nguy{\~{\^e}}n~Th{\d i}, Nguy{\~{\^e}}n Th{\d i}~Minh, Nikaido, Nikolaev,
  Nitisaroj, Nourian, Nurmi, Ojala, Ojha, Ol{\'u}{\`o}kun, Omura, Onwuegbuzia,
  Osenova, {\"O}stling, {\O}vrelid, {\"O}zate{\c s}, {\"O}zg{\"u}r,
  {\"O}zt{\"u}rk~Ba{\c s}aran, Partanen, Pascual, Passarotti, Patejuk,
  Paulino-Passos, Peljak-{\L}api{\'n}ska, Peng, Perez, Perkova, Perrier,
  Petrov, Petrova, Phelan, Piitulainen, Pirinen, Pitler, Plank, Poibeau,
  Ponomareva, Popel, Pretkalni{\c n}a, Pr{\'e}vost, Prokopidis,
  Przepi{\'o}rkowski, Puolakainen, Pyysalo, Qi, R{\"a}{\"a}bis, Rademaker,
  Rama, Ramasamy, Ramisch, Rashel, Rasooli, Ravishankar, Real, Rebeja, Reddy,
  Rehm, Riabov, Rie{\ss}ler, Rimkut{\.e}, Rinaldi, Rituma, Rocha,
  R{\"o}gnvaldsson, Romanenko, Rosa, Roșca, Rovati, Rudina, Rueter,
  R{\'u}narsson, Sadde, Safari, Sagot, Sahala, Saleh, Salomoni, Samard{\v
  z}i{\'c}, Samson, Sanguinetti, S{\"a}rg, Saul{\={\i}}te, Sawanakunanon,
  Scannell, Scarlata, Schneider, Schuster, Seddah, Seeker, Seraji, Shen,
  Shimada, Shirasu, Shohibussirri, Sichinava, Sigurðsson, Silveira, Silveira,
  Simi, Simionescu, Simk{\'o}, {\v S}imkov{\'a}, Simov, Skachedubova, Smith,
  Soares-Bastos, Spadine, Steingr{\'{\i}}msson, Stella, Straka, Strickland,
  Strnadov{\'a}, Suhr, Sulestio, Sulubacak, Suzuki, Sz{\'a}nt{\'o}, Taji,
  Takahashi, Tamburini, Tan, Tanaka, Tella, Tellier, Thomas, Torga, Toska,
  Trosterud, Trukhina, Tsarfaty, T{\"u}rk, Tyers, Uematsu, Untilov, Ure{\v
  s}ov{\'a}, Uria, Uszkoreit, Utka, Vajjala, van Niekerk, van Noord, Varga,
  Villemonte de~la Clergerie, Vincze, Wakasa, Wallenberg, Wallin, Walsh, Wang,
  Washington, Wendt, Widmer, Williams, Wir{\'e}n, Wittern, Woldemariam, Wong,
  Wr{\'o}blewska, Yako, Yamashita, Yamazaki, Yan, Yasuoka, Yavrumyan, Yu, {\v
  Z}abokrtsk{\'y}, Zahra, Zeldes, Zhu, and Zhuravleva}]{ud-v2.7}
Daniel Zeman, Joakim Nivre, Mitchell Abrams, Elia Ackermann, No{\"e}mi Aepli,
  Hamid Aghaei, {\v Z}eljko Agi{\'c}, Amir Ahmadi, Lars Ahrenberg,
  Chika~Kennedy Ajede, Gabriel{\.e} Aleksandravi{\v c}i{\=u}t{\.e}, Ika Alfina,
  Lene Antonsen, Katya Aplonova, Angelina Aquino, Carolina Aragon, Maria~Jesus
  Aranzabe, {\t H}{\'o}runn Arnard{\'o}ttir, Gashaw Arutie, Jessica~Naraiswari
  Arwidarasti, Masayuki Asahara, Luma Ateyah, Furkan Atmaca, Mohammed Attia,
  Aitziber Atutxa, Liesbeth Augustinus, Elena Badmaeva, Keerthana
  Balasubramani, Miguel Ballesteros, Esha Banerjee, Sebastian Bank, Verginica
  Barbu~Mititelu, Victoria Basmov, Colin Batchelor, John Bauer, Seyyit~Talha
  Bedir, Kepa Bengoetxea, G{\"o}zde Berk, Yevgeni Berzak, Irshad~Ahmad Bhat,
  Riyaz~Ahmad Bhat, Erica Biagetti, Eckhard Bick, Agn{\.e} Bielinskien{\.e},
  Krist{\'{\i}}n Bjarnad{\'o}ttir, Rogier Blokland, Victoria Bobicev,
  Lo{\"{\i}}c Boizou, Emanuel Borges~V{\"o}lker, Carl B{\"o}rstell, Cristina
  Bosco, Gosse Bouma, Sam Bowman, Adriane Boyd, Kristina Brokait{\.e}, Aljoscha
  Burchardt, Marie Candito, Bernard Caron, Gauthier Caron, Tatiana Cavalcanti,
  G{\"u}l{\c s}en Cebiro{\u g}lu~Eryi{\u g}it, Flavio~Massimiliano Cecchini,
  Giuseppe G.~A. Celano, Slavom{\'{\i}}r {\v C}{\'e}pl{\"o}, Savas Cetin,
  {\"O}zlem {\c C}etino{\u g}lu, Fabricio Chalub, Ethan Chi, Yongseok Cho,
  Jinho Choi, Jayeol Chun, Alessandra~T. Cignarella, Silvie Cinkov{\'a},
  Aur{\'e}lie Collomb, {\c C}a{\u g}r{\i} {\c C}{\"o}ltekin, Miriam Connor,
  Marine Courtin, Elizabeth Davidson, Marie-Catherine de~Marneffe, Valeria
  de~Paiva, Mehmet~Oguz Derin, Elvis de~Souza, Arantza Diaz~de Ilarraza, Carly
  Dickerson, Arawinda Dinakaramani, Bamba Dione, Peter Dirix, Kaja Dobrovoljc,
  Timothy Dozat, Kira Droganova, Puneet Dwivedi, Hanne Eckhoff, Marhaba Eli,
  Ali Elkahky, Binyam Ephrem, Olga Erina, Toma{\v z} Erjavec, Aline Etienne,
  Wograine Evelyn, Sidney Facundes, Rich{\'a}rd Farkas, Mar{\'{\i}}lia
  Fernanda, Hector Fernandez~Alcalde, Jennifer Foster, Cl{\'a}udia Freitas,
  Kazunori Fujita, Katar{\'{\i}}na Gajdo{\v s}ov{\'a}, Daniel Galbraith, Marcos
  Garcia, Moa G{\"a}rdenfors, Sebastian Garza, Fabr{\'{\i}}cio~Ferraz Gerardi,
  Kim Gerdes, Filip Ginter, Iakes Goenaga, Koldo Gojenola, Memduh
  G{\"o}k{\i}rmak, Yoav Goldberg, Xavier G{\'o}mez~Guinovart, Berta
  Gonz{\'a}lez~Saavedra, Bernadeta Grici{\=u}t{\.e}, Matias Grioni, Lo{\"{\i}}c
  Grobol, Normunds Gr{\= u}z{\={\i}}tis, Bruno Guillaume, C{\'e}line
  Guillot-Barbance, Tunga G{\"u}ng{\"o}r, Nizar Habash, Hinrik Hafsteinsson,
  Jan Haji{\v c}, Jan Haji{\v c}~jr., Mika H{\"a}m{\"a}l{\"a}inen, Linh
  H{\`a}~M{\~y}, Na-Rae Han, Muhammad~Yudistira Hanifmuti, Sam Hardwick, Kim
  Harris, Dag Haug, Johannes Heinecke, Oliver Hellwig, Felix Hennig, Barbora
  Hladk{\'a}, Jaroslava Hlav{\'a}{\v c}ov{\'a}, Florinel Hociung, Petter Hohle,
  Eva Huber, Jena Hwang, Takumi Ikeda, Anton~Karl Ingason, Radu Ion, Elena
  Irimia, {\d O}l{\'a}j{\'{\i}}d{\'e} Ishola, Tom{\'a}{\v s} Jel{\'{\i}}nek,
  Anders Johannsen, Hildur J{\'o}nsd{\'o}ttir, Fredrik J{\o}rgensen, Markus
  Juutinen, Sarveswaran K, H{\"u}ner Ka{\c s}{\i}kara, Andre Kaasen, Nadezhda
  Kabaeva, Sylvain Kahane, Hiroshi Kanayama, Jenna Kanerva, Boris Katz, Tolga
  Kayadelen, Jessica Kenney, V{\'a}clava Kettnerov{\'a}, Jesse Kirchner, Elena
  Klementieva, Arne K{\"o}hn, Abdullatif K{\"o}ksal, Kamil Kopacewicz, Timo
  Korkiakangas, Natalia Kotsyba, Jolanta Kovalevskait{\.e}, Simon Krek,
  Parameswari Krishnamurthy, Sookyoung Kwak, Veronika Laippala, Lucia Lam,
  Lorenzo Lambertino, Tatiana Lando, Septina~Dian Larasati, Alexei Lavrentiev,
  John Lee, Phuong L{\^e}~H{\`{\^o}}ng, Alessandro Lenci, Saran Lertpradit,
  Herman Leung, Maria Levina, Cheuk~Ying Li, Josie Li, Keying Li, Yuan Li,
  {KyungTae} Lim, Krister Lind{\'e}n, Nikola Ljube{\v s}i{\'c}, Olga Loginova,
  Andry Luthfi, Mikko Luukko, Olga Lyashevskaya, Teresa Lynn, Vivien Macketanz,
  Aibek Makazhanov, Michael Mandl, Christopher Manning, Ruli Manurung, C{\u
  a}t{\u a}lina M{\u a}r{\u a}nduc, David Mare{\v c}ek, Katrin Marheinecke,
  H{\'e}ctor Mart{\'{\i}}nez~Alonso, Andr{\'e} Martins, Jan Ma{\v s}ek, Hiroshi
  Matsuda, Yuji Matsumoto, Ryan {McDonald}, Sarah {McGuinness}, Gustavo
  Mendon{\c c}a, Niko Miekka, Karina Mischenkova, Margarita Misirpashayeva,
  Anna Missil{\"a}, C{\u a}t{\u a}lin Mititelu, Maria Mitrofan, Yusuke Miyao,
  {AmirHossein} Mojiri~Foroushani, Amirsaeid Moloodi, Simonetta Montemagni,
  Amir More, Laura Moreno~Romero, Keiko~Sophie Mori, Shinsuke Mori, Tomohiko
  Morioka, Shigeki Moro, Bjartur Mortensen, Bohdan Moskalevskyi, Kadri
  Muischnek, Robert Munro, Yugo Murawaki, Kaili M{\"u}{\"u}risep, Pinkey
  Nainwani, Mariam Nakhl{\'e}, Juan~Ignacio Navarro~Hor{\~n}iacek, Anna
  Nedoluzhko, Gunta Ne{\v s}pore-B{\=e}rzkalne, Luong Nguy{\~{\^e}}n~Th{\d i},
  Huy{\`{\^e}}n Nguy{\~{\^e}}n Th{\d i}~Minh, Yoshihiro Nikaido, Vitaly
  Nikolaev, Rattima Nitisaroj, Alireza Nourian, Hanna Nurmi, Stina Ojala,
  Atul~Kr. Ojha, Ad{\'e}day{\d o} Ol{\'u}{\`o}kun, Mai Omura, Emeka
  Onwuegbuzia, Petya Osenova, Robert {\"O}stling, Lilja {\O}vrelid, {\c
  S}aziye~Bet{\"u}l {\"O}zate{\c s}, Arzucan {\"O}zg{\"u}r, Balk{\i}z
  {\"O}zt{\"u}rk~Ba{\c s}aran, Niko Partanen, Elena Pascual, Marco Passarotti,
  Agnieszka Patejuk, Guilherme Paulino-Passos, Angelika Peljak-{\L}api{\'n}ska,
  Siyao Peng, Cenel-Augusto Perez, Natalia Perkova, Guy Perrier, Slav Petrov,
  Daria Petrova, Jason Phelan, Jussi Piitulainen, Tommi~A Pirinen, Emily
  Pitler, Barbara Plank, Thierry Poibeau, Larisa Ponomareva, Martin Popel,
  Lauma Pretkalni{\c n}a, Sophie Pr{\'e}vost, Prokopis Prokopidis, Adam
  Przepi{\'o}rkowski, Tiina Puolakainen, Sampo Pyysalo, Peng Qi, Andriela
  R{\"a}{\"a}bis, Alexandre Rademaker, Taraka Rama, Loganathan Ramasamy, Carlos
  Ramisch, Fam Rashel, Mohammad~Sadegh Rasooli, Vinit Ravishankar, Livy Real,
  Petru Rebeja, Siva Reddy, Georg Rehm, Ivan Riabov, Michael Rie{\ss}ler, Erika
  Rimkut{\.e}, Larissa Rinaldi, Laura Rituma, Luisa Rocha, Eir{\'{\i}}kur
  R{\"o}gnvaldsson, Mykhailo Romanenko, Rudolf Rosa, Valentin Roșca, Davide
  Rovati, Olga Rudina, Jack Rueter, Kristj{\'a}n R{\'u}narsson, Shoval Sadde,
  Pegah Safari, Beno{\^{\i}}t Sagot, Aleksi Sahala, Shadi Saleh, Alessio
  Salomoni, Tanja Samard{\v z}i{\'c}, Stephanie Samson, Manuela Sanguinetti,
  Dage S{\"a}rg, Baiba Saul{\={\i}}te, Yanin Sawanakunanon, Kevin Scannell,
  Salvatore Scarlata, Nathan Schneider, Sebastian Schuster, Djam{\'e} Seddah,
  Wolfgang Seeker, Mojgan Seraji, Mo~Shen, Atsuko Shimada, Hiroyuki Shirasu,
  Muh Shohibussirri, Dmitry Sichinava, Einar~Freyr Sigurðsson, Aline Silveira,
  Natalia Silveira, Maria Simi, Radu Simionescu, Katalin Simk{\'o}, M{\'a}ria
  {\v S}imkov{\'a}, Kiril Simov, Maria Skachedubova, Aaron Smith, Isabela
  Soares-Bastos, Carolyn Spadine, Stein{\t h}{\'o}r Steingr{\'{\i}}msson,
  Antonio Stella, Milan Straka, Emmett Strickland, Jana Strnadov{\'a}, Alane
  Suhr, Yogi~Lesmana Sulestio, Umut Sulubacak, Shingo Suzuki, Zsolt
  Sz{\'a}nt{\'o}, Dima Taji, Yuta Takahashi, Fabio Tamburini, Mary Ann~C. Tan,
  Takaaki Tanaka, Samson Tella, Isabelle Tellier, Guillaume Thomas, Liisi
  Torga, Marsida Toska, Trond Trosterud, Anna Trukhina, Reut Tsarfaty, Utku
  T{\"u}rk, Francis Tyers, Sumire Uematsu, Roman Untilov, Zde{\v n}ka Ure{\v
  s}ov{\'a}, Larraitz Uria, Hans Uszkoreit, Andrius Utka, Sowmya Vajjala,
  Daniel van Niekerk, Gertjan van Noord, Viktor Varga, Eric Villemonte de~la
  Clergerie, Veronika Vincze, Aya Wakasa, Joel~C. Wallenberg, Lars Wallin,
  Abigail Walsh, Jing~Xian Wang, Jonathan~North Washington, Maximilan Wendt,
  Paul Widmer, Seyi Williams, Mats Wir{\'e}n, Christian Wittern, Tsegay
  Woldemariam, Tak-sum Wong, Alina Wr{\'o}blewska, Mary Yako, Kayo Yamashita,
  Naoki Yamazaki, Chunxiao Yan, Koichi Yasuoka, Marat~M. Yavrumyan, Zhuoran Yu,
  Zden{\v e}k {\v Z}abokrtsk{\'y}, Shorouq Zahra, Amir Zeldes, Hanzhi Zhu, and
  Anna Zhuravleva. 2020.
\newblock \href {http://hdl.handle.net/11234/1-3424} {Universal dependencies
  2.7}.
\newblock {LINDAT}/{CLARIAH}-{CZ} digital library at the Institute of Formal
  and Applied Linguistics ({{\'U}FAL}), Faculty of Mathematics and Physics,
  Charles University.

\end{thebibliography}
\bibliographystyle{acl_natbib}

\vfill\eject

\appendix

\section{Language Selection}
\label{app:languages}

Table \ref{tab:lang-details} shows that the details for languages such as language code, UD treebank id and language family. For POS tagging, we use the Universal Dependencies (UD) 2.7 dataset \cite{ud-v2.7} and for NER, we use WikiANN \citep{pan-etal-2017-cross} with the train, dev and test splits from \citet{rahimi-etal-2019-massively}. To partition languages for the mixed-language multi-task setting, we group languages from the same families into the same partitions to avoid a strong supervision from the same language family when evaluating zero-shot predictions for \textit{unseen} task-language combinations. When there is no available training data in the target treebank, we use the test split for the mixed-language multi-task setting. 

\section{Experimental Details}

\subsection{Impact of Sampling}
\label{sec:sampling}

\begin{figure}[t]
    \vspace{-0.4cm}
    \centering \hspace{-0.4cm}
    \includegraphics[scale=0.195]{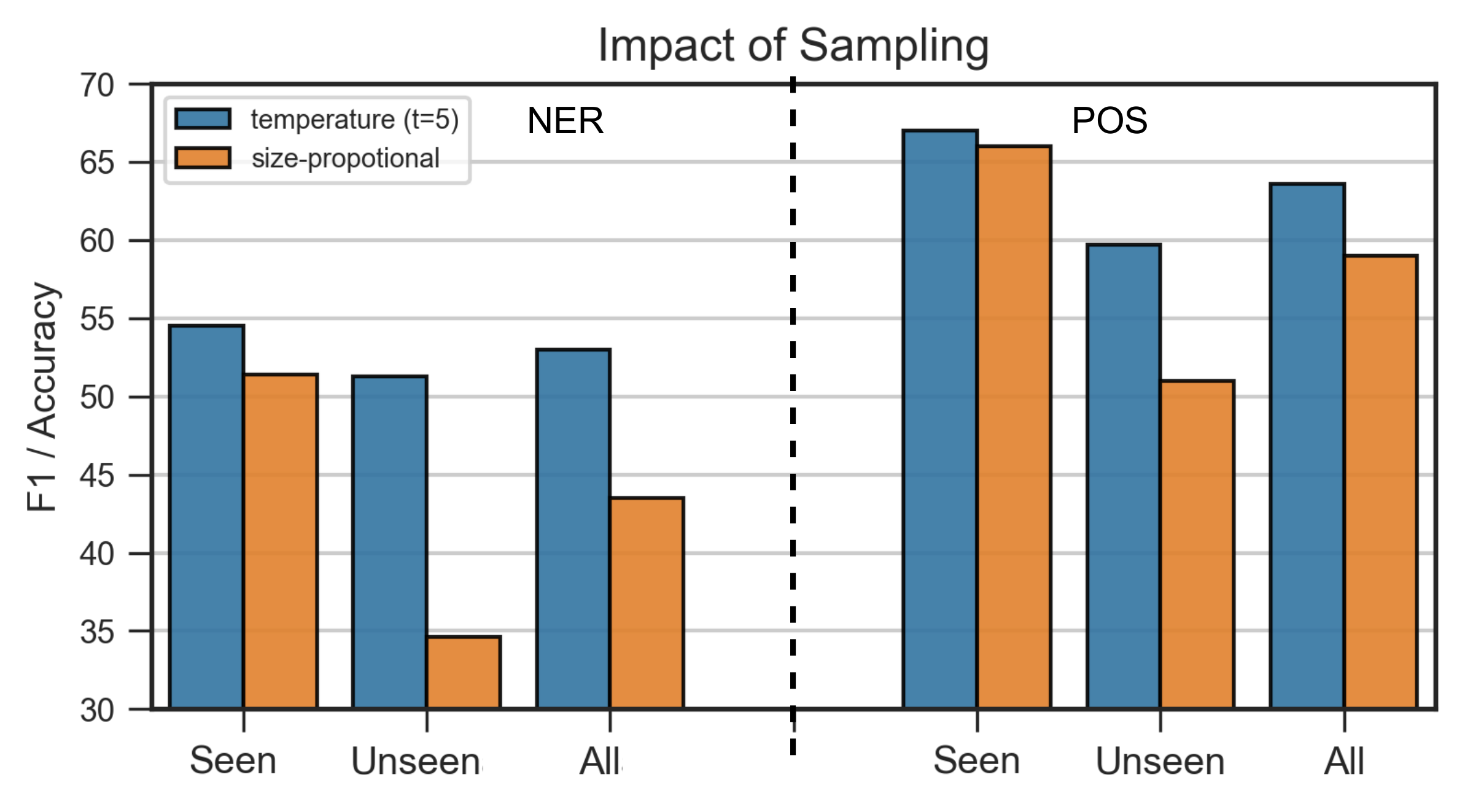}
    \caption{Impact of sampling for \textsc{seen}, \textsc{unseen} language groups on NER and POS tagging.}
  \label{fig:sampling}
\end{figure}

Hyper-X is a single model that is trained at once for multiple languages and task 	simultaneously. However, as the amount of total MLM training data is considearbly larger than NER and POS-tagging data, we experimented with two different sampling methods: size propotional sampling and temperature-based sampling ($t=5$). For the temperature-based sampling, we independently sample a batch for each task-language combination. Figure~\ref{fig:sampling} shows the impact of different sampling methods on the zero-shot performance for `seen', `unseen' language groups together with average over all languages. As seen, temperature-based sampling, greatly increase performance for all language groups on both NER and POS-tagging. This suggest that when MLM data does not restricted by sampling, it highly influences the learning objective which results a catastrophic forgetting on the target tasks. 

\subsection{Implementation and Computing Infrastructure} 
All the experiments are conducted using Tesla V100 GPUs. We did not use parallel training on multiple GPUs, so each experiment was conducted on a single GPU. Parameters that are fine-tuned for each model and total runtime are reported in the section (\S~\ref{sec:efficiency}). We implemented Hyper-X by using Transformers library \citep{wolf-etal-2020-transformers} and the code will be released upon publication. We used adapterhub \citep{pfeiffer2020AdapterHub} for MAD-X, and the original repository for parameter space factorization \citep{ponti-etal-2021-parameter}. Hyper-parameters that are used in experiments are given in the section \ref{sec:experiments}. We did not conduct a hyper-parameter search due to the computational limitations, and used the reference values in most cases: only the dimension for language adapters in MAD-X is changed to match with the same parameter count of Hyper-X. Finally for mBERT, we did a preliminary experiments with learning rate of 1e-4 and 1e-5, and pick the latter one as it produced better performance. 

\begin{table*}[h]
\centering
\small
\begin{tabular}{@{}lllllcccccc@{}}
\toprule
&&&&& \multicolumn{3}{c}{NER} & \multicolumn{3}{c}{POS} \\
Language & Code & UD Treebank  & Family &&  Train & Dev & Test & Train & Dev & Test \\ \midrule
English  & en   & EWT       & IE, Germanic  && 20000 & 10000 & 10000 & 12543 & 2002 & 2077 \\
\noalign{\smallskip} 
\cdashline{1-11}
\noalign{\smallskip}
Arabic   & ar   & PADT      & Afro-Asiatic, Semitic && 20000 & 10000 & 10000 & 6075 & 909 & 680 \\
Breton        & br   & KEB           & IE, Celtic  && 1000 & 1000 & 1000 & - & - & 888 \\
Chinese  & zh   & GSD       & Sino-Tibetan    && 20000 & 10000 & 10000 & 3997 & 500 & 500 \\
Icelandic & is & PUD &  IE, Germanic && 1000 & 1000 & 1000 & - & - & 1000 \\
Kazakh        & kk   & KTB           & Turkic, Northwestern  && 1000 & 1000 & 1000  & 32 & - & 1047 \\
Tamil         & ta   & TTB           & Dravidian, Southern    && 15000 & 1000 & 1000 & 400 & 80 & 120 \\
Turkish  & tr   & IMST      & Turkic, Southwestern && 20000 & 10000 & 10000 & 3664 & 988 & 983 \\ 
Yoruba        & yo   & YTB           & Niger-Congo, Defoid  && 100 & 100 & 100 & - & - & 318 \\
\noalign{\smallskip} 
\cdashline{1-11}
\noalign{\smallskip}
Faroese       & fo   & OFT           & IE, Germanic && 100 & 100 & 100  & - & - & 1208 \\
Guarani     & gn   & Thomas           & Tupian, Tupi-Guarani   && 100 & 100 & 100 & - & - & 98 \\
Upper Sorbian        & yo   & UFAL           & IE, Slavic  && 100 & 100 & 100 & - & - & 23 \\
Maltese & mt & MUDT &  Afro-Asiatic, Semitic  && 100 & 100 & 100  & 1123 & 433 & 518 \\
Sanskrit        & sa   & UFAL           & Indic   && 100 & 100 & 100 & - & - & 230 \\
Uyghur & ug & UDT & Turkic, Southeastern && 100 & 100 & 100 & 1656 & 900 & 900 \\
Cantonese     & yue  & HK            & Sino-Tibetan && 20000 & 10000 & 10000 & - & - & 1004 \\
\bottomrule
\end{tabular}
\caption{Languages that are used in the experiments, together with corresponding language code, UD treebank and language families. We used WikiANN \citep{pan-etal-2017-cross,rahimi-etal-2019-massively} and UD version 2.7 \citep{ud-v2.7} for NER and POS-tagging respectively. }
\label{tab:lang-details}
\end{table*}

\begin{table*}[h]
\setlength{\tabcolsep}{2.8pt}
\small
\centering
\begin{tabular}{@{}l:lllllllllllllllllllllll@{}}
\toprule
& \multicolumn{5}{c}{mBERT} && \multicolumn{5}{c}{MAD-X} && \multicolumn{5}{c}{Hyper-X Small} &&\multicolumn{5}{c}{Hyper-X Base}  \\
\noalign{\smallskip}
\cline{2-6} \cline{8-12} \cline{14-18} \cline{20-24}
\noalign{\smallskip}
   & ar   & tr   & zh   & mt   & ug   &  & ar   & tr   & zh   & mt   & ug   &  & ar   & tr   & zh   & mt   & ug   &  & ar   & tr   & zh   & mt   & ug   \\ \midrule
0  & 42.6 & 72.5 & 36.4 & 43.4 & 12.5 &  & 40.3 & 71.5 & 34.9 & 64.4 & 30.4 &  & 37.2 & 71.6 & 34.2 & 61.3 & 22.4 &  & 39.9 & 73.2 & 34.6 & 63.6 & 22.5 \\
5  & 54.7 & 72.8 & 42.0   & 53.9 & 21.8 &  & 52.4 & 73.5 & 39.3 & 67.3 & 37.5 &  & 56.5 & 74.6 & 39.5 & 66.4 & 28.0   &  & 56.9 & 72.9 & 39.3 & 65.6 & 30.4 \\
10 & 69.2 & 76.0   & 42.1 & 53.4 & 30.4 &  & 64.1 & 75.2 & 43.8 & 76.1 & 44.3 &  & 65.1 & 75.0   & 44.9 & 78.0  & 39.6 &  & 67.3 & 74.2 & 44.4 & 78.3 & 34.2 \\
20 & 69.5 & 78.5 & 49.4 & 53.2 & 30.2 &  & 66.1 & 77.4 & 48.6 & 82.1 & 45.1 &  & 66.8 & 76.7 & 51.9 & 80.3 & 39.8 &  & 68.8 & 77.8 & 52.1 & 80.9 & 43.8 \\
50 & 74.5 & 82.1 & 52.3 & 69.1 & 42.5 &  & 70.2 & 81.0   & 50.7 & 84.9 & 60.6 &  & 71.7 & 80.9 & 54.6 & 82.1 & 53.2 &  & 73.7 & 80.9 & 54.8 & 83.6 & 52.5 \\ \midrule
0  & 53.4 & 72.0 & 67.5 & 24.6 & 28.9 &  & 54.0 & 73.2 & 67.3 & 70.8 & 57.3 &  & 53.4 & 69.2 & 65.6 & 58.8 & 40.4 &  & 54.4 & 71.0 & 66.5 & 59.7 & 50.6 \\
5  & 76.2 & 75.1 & 73.1 & 51.7 & 55.8 &  & 76.4 & 76.3 & 73.3 & 80.1 & 72.4 &  & 75.4 & 75.7 & 76.3 & 73.2 & 62.1 &  & 78.4 & 74.2 & 77.9 & 75.6 & 63.9 \\
10 & 81.8 & 76.6 & 79.5 & 60.8 & 58.9 &  & 83.4 & 76.9 & 78.6 & 83.8 & 73.9 &  & 84.3 & 76.8 & 81.6 & 75.3 & 63.9 &  & 84.8 & 75.9 & 81.9 & 79.3 & 66.0   \\
20 & 86.9 & 78.6 & 84.3 & 68.7 & 60.3 &  & 86.7 & 79.3 & 84.2 & 85.8 & 74.7 &  & 87.2 & 78.4 & 87.1 & 78.9 & 65.9 &  & 87.3 & 76.7 & 86.8 & 82.3 & 67.5 \\
50 & 90.2 & 81.3 & 89.1 & 77.9 & 67.3 &  & 90.5 & 81.9 & 88.4 & 90.1 & 77.2 &  & 90.8 & 82.3 & 90.4 & 83.4 & 66.3 &  & 91.2 & 81.6 & 90.8 & 86.0   & 69.0   \\ 
   \bottomrule
\end{tabular}
\caption{Per language results for few-shot experiments, where models are further fine-tuned with a few training instances (0, 5, 10, 20, 50) from NER and POS datasets. For the language selection, \texttt{ar,tr,zh} are covered by mBERT and \texttt{mt,ug} are unseen.}
\label{tab:few-shot}
\end{table*}

\section{Detailed Results}

The results that are averaged over 3 runs for each language are given in Table \ref{tab:detailed-results}

\begin{table*}[h]
\setlength{\tabcolsep}{4pt}
\small
\centering
\begin{tabular}{llcccccccccccccc}
\toprule
&& \multicolumn{4}{c}{English} && \multicolumn{3}{c}{English} && \multicolumn{5}{c}{Mixed-Language} \\
&& \multicolumn{4}{c}{Single-Task} && \multicolumn{3}{c}{Multi-Task} && \multicolumn{5}{c}{Multi-Task} \\
\noalign{\smallskip}
\cline{3-6} \cline{8-10} \cline{12-16}
\noalign{\smallskip}
&& \texttt{mB} & \texttt{MX} & \texttt{HX.32}  & \texttt{HX.192} && \texttt{mB} & \texttt{HX.32}  & \texttt{HX.192} && \texttt{mB} & \texttt{PSF} & \texttt{MX} & \texttt{HX.32}  & \texttt{HX.192} \\ \midrule
\parbox[t]{2mm}{\multirow{18}{*}{\rotatebox[origin=c]{90}{Named entity recognition}}} &
$\textrm{en}^{a,b}$   & 84.2 & 81.6 & 83.6 & 83.8 &  & 83.6 & 82.1 & 82.6 &  & 81.8 & 79.2 & 82.2 & 83.8 & 83.7 \\
\noalign{\smallskip} 
\cdashline{2-16}
\noalign{\smallskip} 
&$\textrm{ar}^b$ & 40.6  & 40.3 & 42.9      & 39.7       &  & 42.6  & 37.2      & 39.9       &  & 45.5  & 43.4 & 53.5 & 47.8      & 49.2       \\
&$\textrm{br}^a$   & 62.9  & 67.2 & 67.1      & 70.2       &  & 66.5  & 66.5      & 69.5       &  & 70.5  & 70.9 & 72.3 & 74.7      & 76.1       \\
&$\textrm{is}^b$   & 65.0  & 70.0 & 71.0      & 72.9       &  & 69.2  & 70.7      & 73.5       &  & 70.6  & 73.5 & 77.5 & 77.3      & 80.2       \\
&$\textrm{kk}^a$  & 47.2  & 46.7 & 49.6      & 46.3       &  & 45.9  & 42.6      & 47.3       &  & 55.4  & 57.1 & 58.9 & 64.5      & 59.0       \\
&$\textrm{ta}^b$   & 53.8  & 51.0 & 47.3      & 50.6       &  & 50.6  & 49.7      & 51.0       &  & 53.7  & 52.2 & 60.4 & 61.1      & 62.2       \\
&$\textrm{tr}^a$   & 70.2  & 71.5 & 73.8      & 71.4       &  & 72.5  & 71.6      & 72.5       &  & 77.2  & 78.2 & 78.9 & 80.3      & 82.7       \\
&$\textrm{yo}^a$   & 47.6  & 53.0 & 42.1      & 50.2       &  & 46.8  & 44.9      & 47.1       &  & 43.4  & 45.6 & 54.4 & 44.8      & 50.2       \\
&$\textrm{zh}^a$   & 39.5  & 34.9 & 39.5      & 34.7       &  & 36.4  & 34.2      & 34.6       &  & 35.0  & 43.5 & 43.3 & 45.7      & 46.5       \\
\noalign{\smallskip} 
\cdashline{2-16}
\noalign{\smallskip}
&$\textrm{gn}^a$  & 43.6  & 50.3 & 49.0      & 57.5       &  & 41.7  & 55.4      & 54.1       &  & 52.2  & 56.8 & 65.0 & 63.5      & 66.1       \\
&$\textrm{hsb}^b$  & 65.4  & 75.6 & 62.2      & 68.6       &  & 61.4  & 64.3      & 74.6       &  & 73.8  & 75.3 & 84.1 & 78.5      & 80.0       \\
&$\textrm{fo}^b$  & 62.1  & 69.1 & 69.0      & 70.7       &  & 60.7  & 74.7      & 74.9       &  & 63.3  & 68.4 & 83.1 & 76.8      & 82.2       \\
&$\textrm{mt}^b$  & 34.1  & 64.4 & 63.0      & 62.8       &  & 43.4  & 61.3      & 63.6       &  & 61.1  & 73.9 & 73.4 & 67.7      & 77.8       \\
&$\textrm{sa}^a$  & 29.6  & 33.1 & 33.2      & 34.6       &  & 29.0  & 30.3      & 30.8       &  & 30.4  & 43.7 & 48.2 & 43.2      & 43.6       \\
&$\textrm{ug}^b$  & 12.8  & 30.4 & 20.1      & 21.1       &  & 12.5  & 22.4      & 22.5       &  & 23.8  & 16.4 & 38.8 & 33.7      & 27.5       \\
&$\textrm{yue}^a$  & 34.6  & 34.8 & 37.7      & 39.5       &  & 34.3  & 36.6      & 37.3       &  & 36.6  & 44.0 & 42.5 & 44.4      & 49.6       \\
\noalign{\smallskip} 
\cdashline{2-16}
\noalign{\smallskip}
&\textsc{seen}   & 53.4  & 54.3 & 54.2      & 54.5       &  & 53.8  & 52.2      & 54.4       &  & 56.4  & 58.1 & 62.4 & 62.0      & 63.3       \\
&\textsc{unseen} & 40.3  & 51.1 & 47.7      & 50.7       &  & 40.4  & 49.3      & 51.1       &  & 48.7  & 54.1 & 62.2 & 58.3      & 61.0       \\
&\textsc{all}    & 47.3  & 52.8 & 51.2      & 52.7       &  & 47.6  & 50.8      & 52.9       &  & 52.8  & 56.2 & 62.3 & 60.3      & 62.3       \\ \midrule\midrule
\parbox[t]{2mm}{\multirow{18}{*}{\rotatebox[origin=c]{90}{Part-of-speech tagging}}} &
$\textrm{en}^{a,b}$   & 97.0 & 96.8 & 96.6 & 96.1 &  & 96.9 & 96.5 & 96.8 &  & 96.5 & 95.3 & 96.7 & 96.7 & 96.7 \\
\noalign{\smallskip} 
\cdashline{2-16}
\noalign{\smallskip} 
&$\textrm{ar}^a$   & 53.4  & 54.0 & 53.1      & 54.4       &  & 52.6  & 53.4      & 54.4       &  & 62.0  & 67.6 & 55.9 & 61.6      & 65.0       \\
&$\textrm{br}^b$ & 66.8  & 70.5 & 65.2      & 70.8       &  & 68.6  & 69.9      & 70.4       &  & 64.7  & 69.7 & 73.8 & 74.9      & 72.5       \\
&$\textrm{is}^a$   & 82.1  & 82.8 & 83.1      & 83.9       &  & 84.1  & 82.4      & 83.0       &  & 83.2  & 81.6 & 84.7 & 85.4      & 85.8       \\
&$\textrm{kk}^b$   & 74.6  & 75.2 & 73.1      & 75.7       &  & 75.2  & 72.2      & 75.1       &  & 70.4  & 79.7 & 80.6 & 80.4      & 80.5       \\
&$\textrm{ta}^a$  & 58.0  & 59.1 & 58.5      & 59.5       &  & 58.5  & 52.6      & 58.6       &  & 63.1  & 67.2 & 62.2 & 61.7      & 62.7       \\
&$\textrm{tr}^b$    & 72.0  & 73.2 & 70.6      & 70.4       &  & 70.1  & 69.2      & 71.0       &  & 70.6  & 73.5 & 75.1 & 74.8      & 75.6       \\
&$\textrm{yo}^b$   & 55.6  & 60.3 & 58.3      & 60.0       &  & 58.4  & 55.2      & 56.6       &  & 58.8  & 57.4 & 64.2 & 61.0      & 63.2       \\
&$\textrm{zh}^b$   & 67.5  & 67.3 & 70.2      & 67.4       &  & 63.1  & 65.6      & 66.5       &  & 64.9  & 66.6 & 69.2 & 65.8      & 66.8       \\
\noalign{\smallskip} 
\cdashline{2-16}
\noalign{\smallskip}
&$\textrm{gn}^b$  & 27.2  & 34.9 & 31.2      & 37.0       &  & 28.3  & 35.1      & 36.7       &  & 38.6  & 36.3 & 44.5 & 40.8      & 41.1       \\
&$\textrm{hsb}^a$  & 71.3  & 76.2 & 75.7      & 73.9       &  & 69.9  & 75.3      & 73.2       &  & 70.3  & 69.0 & 80.4 & 77.5      & 78.5       \\
&$\textrm{fo}^a$  & 87.2  & 88.3 & 86.4      & 87.9       &  & 80.5  & 85.8  & 86.4       &  & 82.1  & 81.1 & 88.9 & 88.6      & 88.6       \\
&$\textrm{mt}^a$  & 24.6  & 70.8 & 61.4      & 52.7       &  & 28.2  & 58.8      & 59.7       &  & 40.7  & 38.1 & 74.3 & 63.9      & 64.0       \\
&$\textrm{sa}^b$  & 39.4  & 46.3 & 43.1      & 39.5       &  & 40.5  & 46.3      & 45.9       &  & 48.1  & 50.4 & 54.5 & 56.6      & 54.6       \\
&$\textrm{ug}^a$  & 28.9  & 57.3 & 44.3      & 56.4       &  & 26.7  & 40.4      & 50.6       &  & 40.2  & 37.2 & 59.7 & 53.0      & 56.0       \\
&$\textrm{yue}^b$  & 63.6  & 64.2 & 62.9      & 63.6       &  & 63.1  & 62.4      & 64.0       &  & 63.2  & 64.6 & 66.4 & 62.2      & 64.0       \\
\noalign{\smallskip} 
\cdashline{2-16}
\noalign{\smallskip}
&\textsc{seen}   & 66.3  & 67.7 & 66.5      & 67.8       &  & 66.3  & 65.1      & 67.0       &  & 67.2  & 70.4 & 70.7 & 70.7      & 71.5       \\
&\textsc{unseen} & 48.9  & 62.6 & 57.9      & 58.7       &  & 48.2  & 57.7      & 59.5       &  & 54.7  & 53.8 & 67.0 & 63.2      & 63.8       \\
&\textsc{all}    & 58.1  & 65.4 & 62.5      & 63.5       &  & 57.9  & 61.7      & 63.6       &  & 61.4  & 62.7 & 69.0 & 67.2      & 67.9    \\   \bottomrule
\end{tabular}
\caption{Zero-shot cross-lingual transfer results averaged over 3 runs for Named-Entity Recognition (NER; F1) and Part-of-Speech Tagging (POS; Accuracy) for 
mBERT (mB), MAD-X (MX) and parameter space factorization (PSF) models, together with Hyper-X Small (HX.32) and Base (HX.192). 
Superscripts denote the partitioning that is used for mixed-language multi-task setting}
\label{tab:detailed-results}
\end{table*}

\section{Few Shot Experiments}
\label{sec:few-shot-exp}
For the few-shot transfer experiments, we fine-tune each model for 50 epochs with the same hyper-parameters. We disable the learning rate decay as only a few training instances are provided to the models. Note that, in these experiments, we always start with the models that are already trained in the zero-shot setting and perform fine-tuning for each language and task separately. For the selection of training samples, we randomly sample instances regardless of the labels, as the initial models are already trained for these tasks on English data.

Table \ref{tab:few-shot} show that few-shot results for NER and POS-tagging respectively. 

\end{document}